\documentclass[twoside,11pt]{article}
\usepackage{jair, theapa, rawfonts}
\usepackage{graphicx}
\usepackage{subcaption}
\usepackage{url}
\usepackage{verbatim}
\usepackage{booktabs}

\jairheading{82}{2025}{2233-2278}{06/2024}{04/2025} 
\ShortHeadings{Detecting AI-Generated Text}
{Fraser, Dawkins \& Kiritchenko}
\firstpageno{2233}

\begin{document}

\title{Detecting AI-Generated Text: Factors Influencing Detectability with Current Methods}

\author{\name Kathleen C. Fraser \email kathleen.fraser@nrc-cnrc.gc.ca \\
       \name Hillary Dawkins \email hillary.dawkins@nrc-cnrc.gc.ca \\      \name Svetlana Kiritchenko \email svetlana.kiritchenko@nrc-cnrc.gc.ca \\
       \addr National Research Council Canada\\
       1200 Montreal Road, Ottawa, Canada
}


\maketitle

\begin{abstract}
Large language models (LLMs) have advanced to a point that even humans have difficulty discerning whether a text was generated by another human, or by a computer. However, knowing whether a text was produced by human or artificial intelligence (AI) is important to determining its trustworthiness, and has applications in many domains including detecting  fraud and academic dishonesty, as well as combating the spread of misinformation and political propaganda. The task of AI-generated text (AIGT) detection is therefore both very challenging, and highly critical. In this survey, we summarize state-of-the art approaches to AIGT detection, including watermarking, statistical and stylistic analysis, and machine learning classification. We also provide information about existing datasets for this task. Synthesizing the research findings, we aim to provide insight into the salient factors that combine to determine how ``detectable'' AIGT text is under different scenarios, and to make practical recommendations for future work towards this significant technical and societal challenge. 
\end{abstract}



\section{Introduction}\label{sec:intro}

\begin{figure}[t!]
    \centering
    \includegraphics[width=\textwidth]{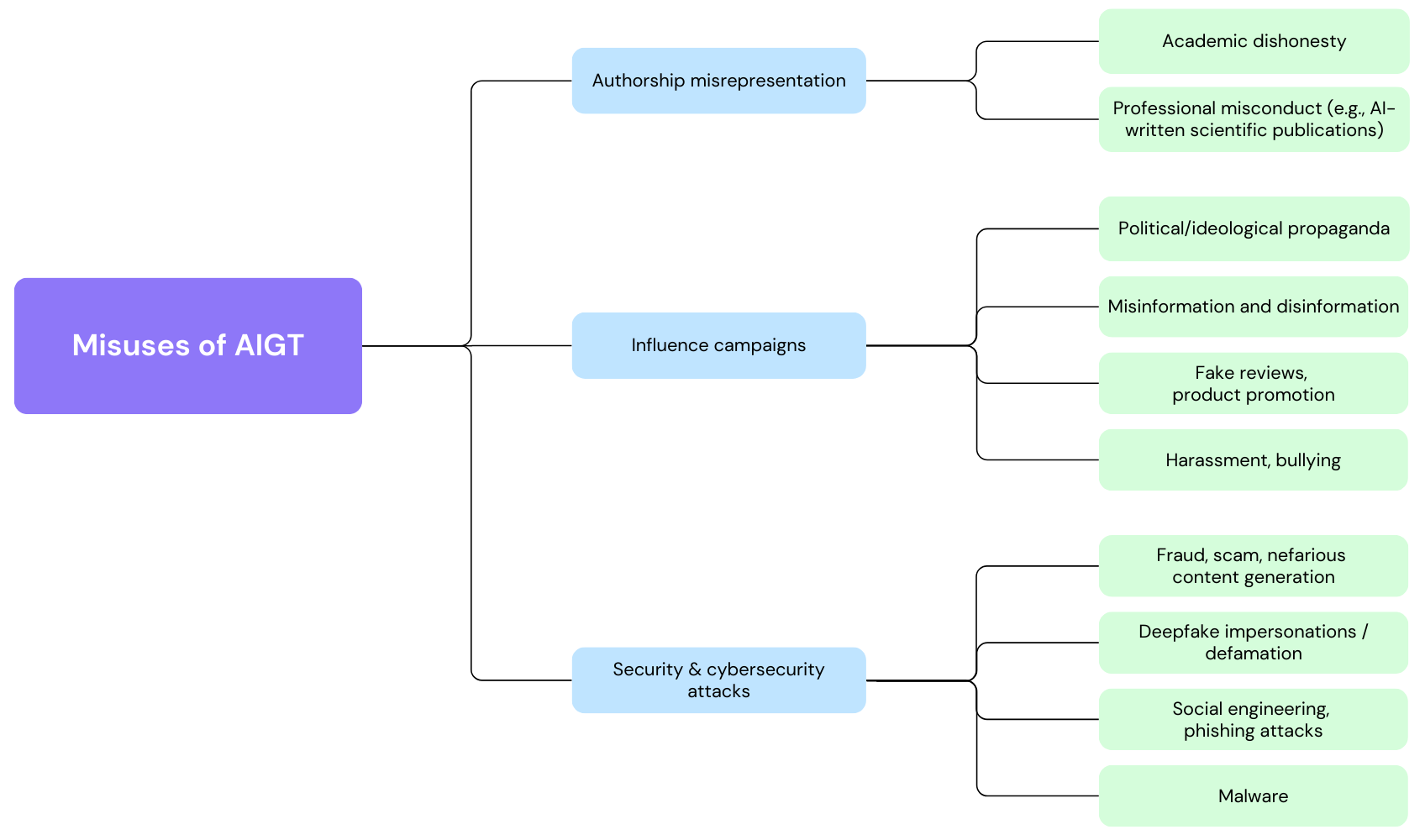}
    \caption{Some common types of misuse of AI-generated text 
    \shortcite{crothers2023machine,weidinger2021ethical}.}
    \label{fig:AIGT_risks}
\end{figure}

In recent years, the capabilities of large language models (LLMs) to generate fluent, realistic-sounding text have improved dramatically. We are now at a point where humans themselves cannot reliably distinguish between text that was generated using artificial intelligence (AI) and that written by a real person \shortcite{liu2023check,sarvazyan2023overview,li2023deepfake}. There are many opportunities for LLMs to contribute to human productivity, with applications to question-answering, computer programming, brainstorming, proof-reading, and information retrieval. 
However, LLMs can also facilitate malicious activities,  increasing efficiency and reducing costs in malware creation, fraud, identity theft, harassment attacks, and academic dishonesty \shortcite{crothers2023machine,wu2023survey} (see Figure~\ref{fig:AIGT_risks}). 
Another major risk brought by generative AI is the potential for global-scale ``information pollution.'' 
Automatically generated `fake' texts can be exploited for commercial goals, such as product promotion or fake product reviews, or for political gains, encompassing propaganda, `fake news', disinformation, etc. 
Current state-of-the-art generative models can produce high-quality, fluent fake information that is perceived as more credible and trustworthy than human-generated misinformation \shortcite{zellers2019defending,spitale2023ai}, and that is harder for both human readers and automatic detection systems to recognize \shortcite{kreps2022all,zhou2023synthetic,chen2023can}. 

To mitigate the risks of malicious use, and to protect the integrity of the information ecosystem, it is essential to develop tools to distinguish between text that was written by a human, and AI-generated text (AIGT). The task of AIGT detection is a challenging problem with constantly moving goalposts: as researchers develop effective methods to detect text from the currently available LLMs, newer and larger models are released and the cycle continues. Furthermore, bad actors seeking to obfuscate their use of AI tools concurrently develop adversarial attacks on the detection methods, aiming to modify their AIGT to render it undetectable to the current methods \shortcite{ghosal2023towards}.  

The present manuscript provides an overview of the field of AIGT detection, the current state-of-the-art methodologies, the available data resources and online tools, and existing challenges. 
In this survey, we have included papers published in both peer-reviewed journals and conference proceedings, as well as those posted on pre-print servers such as arXiv. 
While pre-print manuscripts may not have been formally peer-reviewed yet, we include them in an attempt to cover the most cutting-edge advancements in a rapidly changing field. 
We have focused our attention primarily on papers published in 2023 and 2024, with some references to seminal papers published earlier.\footnote{The cut-off date for this survey was 10 June, 2024. However, we have updated the reference information for papers that appeared as pre-prints before the cut-off and were subsequently published.} We include only those papers that focus on NLP methods for AIGT detection, excluding methods that rely on other characteristics of bot accounts, such as posting frequency or social network analysis. However, within the realm of NLP, we cast our net widely, including methods from diverse data domains (social media, news stories, academic essays, scientific abstracts) and aiming to provide an extensive and high-level overview of the field as it currently stands. 

Other surveys on AIGT detection exist, and we point the interested reader to them for more information. \shortciteA{liu2023survey} and \shortciteA{zhang2023watermarks} focus specifically on watermarking for LLMs. \shortciteA{crothers2023machine} provide an extensive analysis of the risks of AI-generated text, and an evaluation of methods to detect text generated by pre-ChatGPT models. \shortciteA{ghosal2023towards} present an overview of both the \textit{possibilities} (detection methods) and \textit{impossibilities} (evasion methods) of AIGT detection. 
\shortciteA{uchendu2023survey} focus more specifically on the tasks of \textit{author attribution} and \textit{author obfuscation} involving multiple human or AI authors and their combination. 
\shortciteA{tang2023science} offer an accessible summary of major insights and challenges in the field of AIGT detection. Two recent 
surveys and their associated github pages are also useful resources: \shortciteA{yang2023survey}\footnote{\url{https://github.com/Xianjun-Yang/Awesome_papers_on_LLMs_detection}} and \shortciteA{wu2023survey}\footnote{\url{https://github.com/NLP2CT/LLM-generated-Text-Detection}}. In the current work, we hope to build on existing research to achieve the following goals: (1) To summarize emerging and state-of-the-art methods in a rapidly changing field; and (2) To provide a more practical perspective on the problem, by examining the factors that contribute to detectability in any specific use case, taking into account that not all AIGT is equivalent, and that users of detection software are often working with incomplete knowledge of the model that generated the text, the prompt that was input to the model, the underlying decoding strategy, and any steps that may have been taken to obfuscate the provenance of the text. As such, we hope to produce an approachable and useful document for AI practitioners as well as researchers, by highlighting how different features of AI-generated text can impact the performance of different detection algorithms. 











We begin by defining the task of AIGT detection and discussing its key characteristics (Section~\ref{sec:background}).  
In particular, we emphasize that AIGT is not a homogeneous category; it can be thought of as a spectrum from being generated fully automatically, to text with a high degree of human influence (e.g., a document authored by a combination of human and AI, or an AI translation of a human text). 
We also specify several distinct detection scenarios, 
to better inform a comparative analysis of the detection algorithms and their assumptions.

We then outline the current NLP-based approaches to AIGT detection (Section~\ref{sec:approaches}). These are divided into three categories. The first category is watermarking, which involves the creator of the LLM encoding an undetectable signal into the output text, such that anyone who has the watermark detection algorithm can determine that the text came from that model. While fairly straightforward in image generation, watermarking text such that the watermark is both undetectable and does not change the meaning of the sentences is a highly challenging problem. We then discuss approaches based on the observation that LLM language has different statistical and/or linguistic properties than human writing, even if those properties are not always perceptible to human observers. Many of these methods leverage the underlying principle that when an LLM constructs a sentence, it always selects a highly-probable word to come next, given the context, while humans are much more variable in their choices. Therefore, metrics of the statistical regularity of the texts can offer some insight into the source. Finally, we also describe work using pre-trained language models as the basis for the text classification. This approach does not require a feature extraction step, but rather learns directly from examples of text from humans and AI.

We present a (non-exhaustive) list of currently available datasets that include both human-written and AI-generated texts and that could be used for training and/or testing AIGT detection systems (Section~\ref{sec:datasets}). 
Datasets differ along several important axes, including domain (e.g., news, social media, academic writing), language, and generating model settings. In many cases, generalizability across these different parameters appears to be low, so selecting the appropriate dataset for a particular application is crucial. 

Following from our discussion of methods and datasets, we outline some of the different factors that affect how easily detectable a given sample of AIGT is (Section~\ref{sec:factors}). 
We discuss such factors as properties of the generating model (model size and decoding strategy), language of the text, document length, in-distribution vs.\@ out-of-distribution inputs, 
degree of human influence, and adversarial strategies. 
A thorough characterization of these different factors will help enable a user to select the most appropriate detection method and training dataset for a given scenario. 


Finally, 
we summarize the research findings and offer high-level recommendations when designing a solution for a particular application (Section~\ref{sec:discussion}).  
We conclude by highlighting the existing challenges and most promising directions for future work (Section~\ref{sec:conclusion}). As LLMs become even more ubiquitous in our lives, and their power and fluency continue to increase, the detection of AIGT will be a difficult yet critical problem on which researchers, governments, and companies will need to collaborate.

\section{The Task of AIGT Detection}
\label{sec:background}


We begin by defining the task of AIGT detection and providing background information on the relevant text classification and generative AI concepts that will be referenced throughout the survey (Section~\ref{sec:background:NLP}). Additionally, we outline different types of AIGT along a spectrum from minimal to maximal human influence (Section~\ref{sec:taxonomy}), and also present several common detection scenarios and their salient differences, such as whether the detector has knowledge of or access to the original generating model (Section~\ref{sec:background:detection_scenarios}). 


\subsection{Text Classification and Generative AI}
\label{sec:background:NLP}

AIGT detection is a \textit{text classification task}, meaning that the input is a text sequence and the output is a discrete class prediction. 
AIGT detection is usually treated as a binary class problem (``AI'' or ``human''), but could be multi-class if we wish to differentiate the level of AI influence or predict the specific AI model that generated the text (also referred to as an \textit{authorship attribution} task). 
In some cases, a document may be written by a combination of human and AI, and the task is to determine where the boundaries between those sections are. This can also be treated as a text classification problem, by performing the classification on the sentence or paragraph level, and then locating the position where the text switches from one class to the other.

In most cases, AIGT detection is approached using a supervised learning framework, which assumes that labelled examples are available to train or calibrate the classifier. As we will see in the following sections, some detectors are trained using the classical approach of first extracting relevant features from the text (e.g., syntactic or stylistic features) and then feeding those features to a machine learning classifier, either statistical (logistic regression, SVM) or neural (deep neural network).  Other approaches leverage \textit{pre-trained language models} (e.g., BERT, \shortciteA{devlin2018bert}), which are pre-trained in an unsupervised fashion on large text corpora to learn effective representations of the semantic meaning of text as a series of dense distributions. These models can then be fine-tuned for any number of NLP tasks (including AIGT detection), without the need for an explicit feature extraction step.

\begin{figure}[tbh]
     \centering
     \begin{subfigure}[b]{\textwidth}
         \centering
         \includegraphics[width=\textwidth]{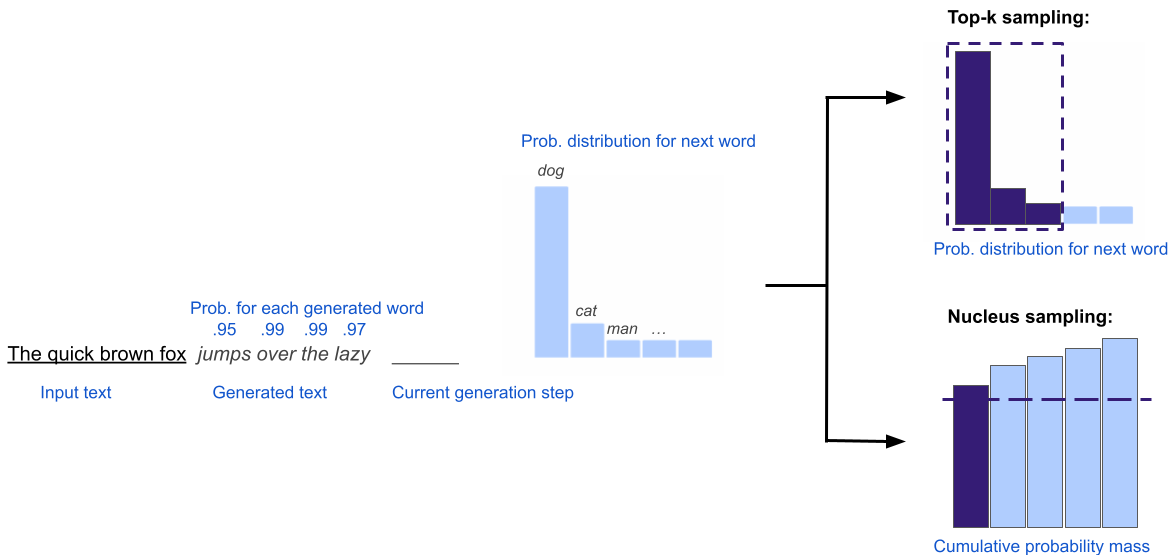}
         \caption{Highly predictable text generation}
         \label{fig:predictable_text_sampling}
     \end{subfigure}
     \vfill
     \begin{subfigure}[b]{\textwidth}
         \centering
         \includegraphics[width=\textwidth]{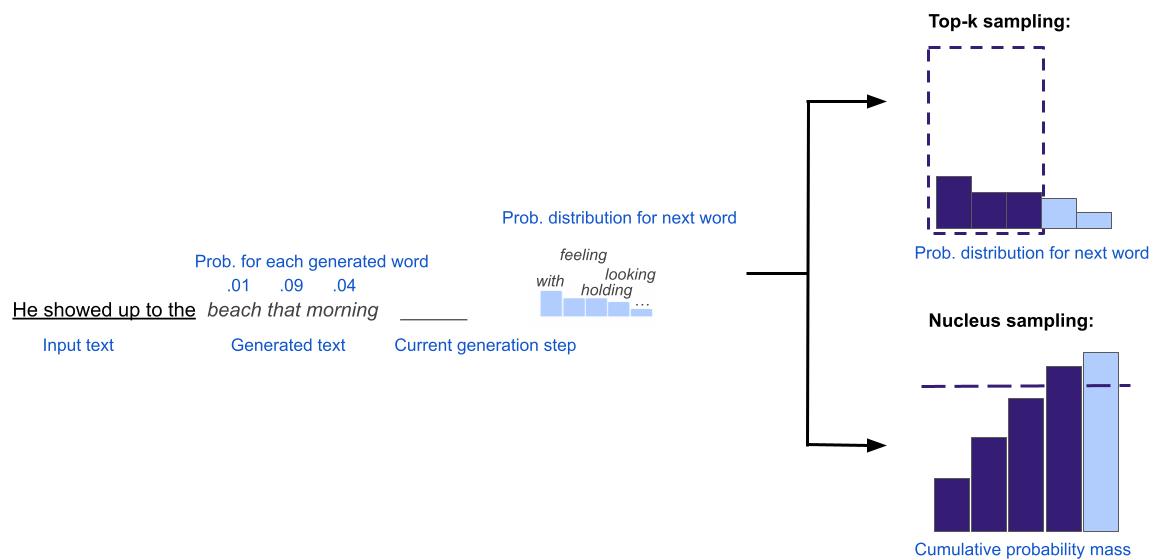}
         \caption{Uncertain text generation}
         \label{fig:usual_text_sampling}
     \end{subfigure}
        \caption{The text is generated one word at a time given the input text by sampling the probability distribution over the vocabulary. (a) When sentences are highly predictable, the probability associated with each generated word is high, and the model is certain about the next generation. (b) More typically, even simple sentences have many reasonable continuations at each generation step, and each possibility takes the story in a different direction.}
        \label{fig:text_sampling}
\end{figure}

While pretrained language models like BERT are trained using a bidirectional framework to predict a masked token, a \textit{generative language model} is trained to predict the next word in a sequence, conditioned on the previous words (i.e., the \textit{context}). 
The early iterations of generative language models (e.g., GPT by \shortciteA{radford2018gpt}) can be thought of simply as pre-trained language models that 
can generate a full continuation of a text by iteratively generating the next word, as shown in Figure~\ref{fig:text_sampling}.
Such models have become known as \textit{large language models}, or LLMs.

In recent years, LLMs have been enhanced through instruction tuning \shortcite{wei2021finetuned} and reinforcement learning with human feedback \shortcite{long2022instruct} to be able to answer questions and follow complex instructions (e.g., GPT-3.5 by \shortciteA{openai2022chatgpt}). 
For example, instead of writing the rest of a story given the first few sentences, an instruction-tuned LLM can write a full story given the \textit{prompt} ``Write a fantasy story in a lighthearted tone.'' The generation still works as it did before, by generating the next word one at a time, but the content and style of the generation are guided by the underlying instruction. 
Note that LLMs can also be used as classifiers, simply by asking the model in plain language to make a prediction. This can be done without any input examples (\textit{zero-shot}), or with a few examples included in the prompt (\textit{few-shot}).    

Nonetheless, the basic functionality of the LLM is still to produce a sequence of words, given the context.  
At each step in the generation process, the LLM has a probability distribution over all words in its vocabulary.
The specific process by which a word is selected from that probability distribution is called the \textit{decoding method}. A \textit{greedy} decoding strategy always selects the highest probability word, but this produces deterministic and repetitive generations. 
In contrast, if the model simply samples the full vocabulary according to the distribution, a poor word choice is likely due to the combined probability mass of all low-probability words. Therefore, a method is needed to choose words that have a high probability given the context (to ensure fluency and consistency) while also sampling from a relatively wide range (to ensure variety and creativity). In practice, two of the most popular decoding strategies are \textit{nucleus sampling} and \textit{top-k sampling}; words are sampled according to their probability values, but the choices are still restricted to only the most probable words. Top-$k$ sampling always selects among the top $k$ most probable choices, whereas nucleus sampling uses the total probability sum as the cut-off criterion, meaning that more choices are possible when the model is uncertain (see Figure~\ref{fig:text_sampling}).
In Section \ref{sec:factors:LLM}, we discuss how the decoding strategy can affect the detectability of the generated text. 

Both the probability values of the selected words and the shape of the probability distribution at each generation step offer important information about how ``surprising'' the text is, or how uncertain the model is (refer to Figure~\ref{fig:text_sampling}). In the NLP literature, there are two commonly used metrics to quantify this uncertainty. 
\textit{Perplexity} is related to the inverse of the length-normalized word sequence probability. For example, if every word in a sentence was predicted with a high probability, the sentence is not very surprising, and it has low perplexity (e.g., Figure~\ref{fig:predictable_text_sampling}).
\textit{Entropy} is a measure of uncertainty or spread in the probability distribution. Highly uniform distributions have high entropy and imply that the model is uncertain about the next word given the context (e.g., Figure~\ref{fig:usual_text_sampling}). 
In Section \ref{sec:statistical}, we will see how these statistical properties of language generation can be used to detect AIGT. Because LLMs generate text by always selecting the next word from the set of highest-probability words, AIGT tends to have low perplexity and low entropy, compared to human texts, as measured by the generating model's probability distribution. 

Throughout the survey, the performance of a classifier/detector will most usually be reported simply by prediction accuracy (i.e., the percentage of text instances for which the label was predicted correctly), with some exceptions. For classifiers that use a flexible decision threshold (more on this in Section \ref{sec:statistical}), it makes sense to look at the area under the receiver operating characteristic (AUROC) curve \shortcite{hajian2013receiver} to observe the average performance over all possible decision thresholds. Note that the flexibility in choosing a decision threshold means that we can choose a fixed false positive rate. In the context of AIGT, we might desire a very low false positive rate (i.e., a low rate of human-written texts being misclassified as AI) to mitigate the repercussions for human users. For this reason, some of the surveyed papers prefer to report performance as the true positive rate at a fixed false positive rate (TPR@FPR) instead of using the average captured by AUROC.

\subsection{Taxonomy of AI-Generated Text}
\label{sec:taxonomy}

One aspect of the task that quickly becomes apparent is that the phrase ``AI-generated text'' covers a wide variety of texts, with differing levels of human input. It can include texts where the content and structure are entirely determined by the AI, such as in response to a prompt like \textit{Tell me a bedtime story}. It can also include texts where the semantic content is specified, but the style and syntax are determined by the AI, as in summarization or paraphrasing. Some researchers have argued that the term AIGT should also cover cases such as machine translation, where the content and structure are highly specified by the original human text, but its final, translated form has been generated by an AI model via machine translation. Finally, we must also consider the various cases where a text is authored by a human but ``polished'' by a language model, or generated by AI but then post-edited by a human, as well as documents in which part of the text was authored by a human and part by AI. These different types of AIGT represent different challenges in terms of detectability; furthermore,  depending on the application, users may want to detect particular categories of AIGT but not others. 

\begin{table}[t!]
    \centering
    \begin{tabular}{l | p{12cm}}
    \hline
    Class & Examples \\
    \hline
    Arbitrary     &  
    \textbf{Totally arbitrary: }
    \textit{Please write a story.}\\
    \hline
    
    Guided     &  \textbf{Topic-guided:} \textit{Given the following headline please write a story. Headline: PFIZER RELEASES NEW COVID VACCINE.} \\
    & \textbf{Message-guided} \textit{Given the following headline please write a story that describes the hidden dangers of vaccines. Headline: PFIZER RELEASES NEW COVID VACCINE.} \\
    \hline
    
    Controlled &\textbf{Paraphrase:} \textit{Given the following text, please re-write it using different words.}\\
    &\textbf{Style transfer:} \textit{Given the following text, re-write it using a more formal tone.} \\
    & \textbf{Summary: }\textit{Given the following text, please summarize it into 2-3 sentences.} \\
    & \textbf{Translation:}  \textit{Given the following story, please translate it into French.} \\
   &  \textbf{Polishing:} \textit{Given the following text, please make small changes for readability, keeping most of the content intact.} \\
    
    \hline 
    
    Collaborative 
    & \textbf{Post-editing:}  Given an AIGT, a human can review and make changes or correct errors. \\
    & \textbf{Mixcase:}  Within a document, some sections are written by humans and some by AI. \\
    & \textbf{Cyborg account:}  Social media accounts which are run partially by bots and partially by human authors. \\
    
    \hline 
    \end{tabular}
    \caption{Categorization of different kinds of AI-generated text, based on the level of human intervention. }
    \label{tab:aigt_types}
\end{table}

Table~\ref{tab:aigt_types} summarizes the different types of AIGT and categorizes them into four high-level classes of generation: Arbitrary, Guided, Controlled, and Collaborative. These categories are based on similar taxonomies in the literature \shortcite{chen2023can,crothers2023machine}, although it should be noted that the boundaries between classes are somewhat fuzzy. 

In \textit{arbitrary} generation, the AI model has the highest degree of freedom to determine both the content and structure of the generated text. Although the user may provide some direction in their prompt, the content will be strongly influenced by the language model training. This type of text generation may be most useful for entertainment, creative tasks, and brainstorming. 

In practice, users are likely to want to convey a more specific intent to the AI system. In \textit{guided} generation, the user indicates to the model the general message or idea that should be conveyed in the output text. Guided generation is potentially dangerous as it allows the user to easily generate a large quantity of text from only a few words of prompting, for example to generate highly-convincing misinformation \shortcite{spitale2023ai} or long essays \shortcite{liu2023argugpt}. 

Even more human control is exercised in \textit{controlled} generation. Here, the complete content of the text is specified, but the language model is used to modify the text in some way: paraphrase, change the style (perhaps make it more professional, or more casual), summarize/shorten, or translate to a different language. This method of producing texts can be extremely powerful, but necessitates the existence of an input text (presumably written by a human, but possibly also itself AI generated). On the boundary between the \textit{controlled} category and the next category we have ``polishing,'' a use case in which a human author inputs their own writing but uses AI to make minor changes, e.g., to improve fluency or readability. This use of AI is still under debate in academic and professional settings, as AI proof-reading tools can help writers improve the communication of their ideas in their second language, but may also be detected by plagiarism/AIGT-detection systems \shortcite{liu2023check}. 

The final category that we consider is \textit{collaborative} generation, where the text is authored by some combination of human and machine \shortcite{cutler2021automatic}. This could involve human-editing of AI text, or discrete sections within a document written by human and AI (defined as `mixcase' by \shortciteA{gao2024llm}). Also falling within this category is the phenomenon of `cyborg' social media accounts, in which a bot generates most of the initial content, but a human takes over to handle replies and individual conversations. If all the posts from that account are collected and concatenated, it forms a special case of mixcase. 

As we will see in the following sections, these four categories are generally progressively harder for detection methods to tackle. Watermarks, statistical regularities, and other features of AI-written text can be weakened or removed through human editing and intervention. At the same time, there is an increasing human cost in terms of time and effort as we progress through the categories, which may motivate users aiming to produce fully-automated content at scale to use primarily arbitrary or guided methods of generation.

\subsection{Detection Scenarios}
\label{sec:background:detection_scenarios}

As the user or designer of a detection system, we will not typically know to which category in Table~\ref{tab:aigt_types} a given text belongs (perhaps with some exceptions: e.g., an AIGT detector used by a university may assume that most problematic cases have used guided generation to produce an essay on the assigned topic, although it is possible that students would use translation, polishing, post-editing, or mixcase as well). However, and separate from that issue, we may also have varying degrees of information about the AI model that was used to generate the text. Different detection methods may be more appropriate depending on our knowledge of, and access to, the generating model. Here we briefly introduce some terminology that will be used as we discuss the algorithms in Section~\ref{sec:approaches}.

\begin{itemize}
    \item \textbf{Known-model scenario:} In this scenario, we know which LLM generated the text we are trying to detect. For example, a company that develops an LLM might also want to train a detector for that specific LLM. In a broader definition of this scenario, we may assume that the set of all possible generating models is known. That is, we know the text came from one of $n$ models.
    \item \textbf{Unknown-model scenario:} In this scenario, you do not know anything about which model generated the text. This is a more challenging scenario, but probably the most realistic for detecting AIGT online. 
    \item \textbf{White-box access:} Here it is assumed that you have either full access to the generating model's entire model weights or at least the output probability values. Clearly, this is only possible in the known-model scenario, and is more common for open-access models. 
    \item \textbf{Black-box access:} Here it is assumed that you can input prompts to a (known) model and observe the outputs, but you do not have information about the internal weights or output probabilities. This prohibits the use of any algorithms which rely on those probability values for detection. OpenAI's GPT-4 is an example where most users have only black-box access to the model.    
\end{itemize}

In the most open-ended detection scenario, we do not know which model was used to generate the text and therefore cannot have any access to its parameters nor generate training data from it, and must rely on proxy models. However, in many real-life scenarios such as academic misconduct, it can be safely assumed that most students will use one of the dozen or so widely-available LLMs, and therefore the problem can be reduced to a series of known-model problems, where the detector checks if the text was generated by any one of the candidate LLMs. Where possible, calibrating the detection methods to have low false positive rates is helpful when using an ensemble of known-model detection methods applied to the unknown-model task. 



\section{Current Approaches to AIGT Detection}\label{sec:approaches}

In the following section, we summarize the existing NLP approaches to AIGT detection, based on our survey of the literature. The methods fall broadly into three high-level categories: watermarking, statistical and stylistic analysis, and using pre-trained language models for classification. Each of these categories is associated with its own strengths and weaknesses, which we highlight as we go through.  

\subsection{Watermarking}
\label{sec:watermarking}

\begin{figure}[t!]
    \centering
    \includegraphics[width=0.9\textwidth]{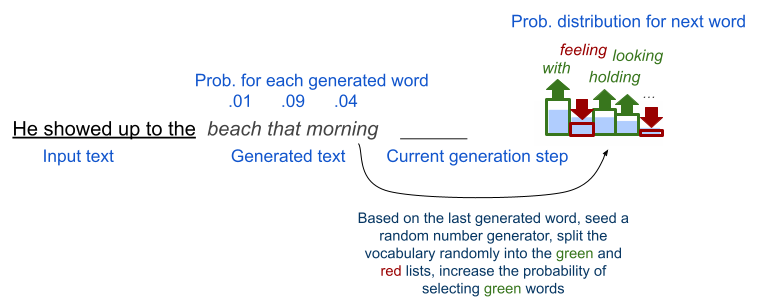}
    \caption{Generating text with watermarking using red-green lists \shortcite{kirchenbauer2023watermark}.}
    \label{fig:watermarking}
\end{figure}

A watermark is an identifier that is secretly embedded in text and used to convey some meta-information. 
In the context of AIGT detection, the idea is that the presence of a watermark denotes that the text is AI-generated. With knowledge of the watermark extraction algorithm, one can verify the text authorship. 
In other contexts, multi-bit watermarking \shortcite{yoo2023robust,yoo-etal-2024-advancing} may be used to encode additional copyright information.
Ideally, inserting a watermark does not alter the meaning or quality of the original text, and it should be imperceptible to those without knowledge of the watermarking method. If an adversary detects the use of a watermark, they might change text generation methods or attempt to erase it. A good watermark should therefore be robust to adversarial attacks such as text perturbations and paraphrasing. 

\shortciteA{kirchenbauer2023watermark} introduce the idea of embedding a watermark during text generation by upsampling specific words in the vocabulary (Figure~\ref{fig:watermarking}). Specifically, at every token location, the vocabulary is divided into a green list and a red list using a random seed generated from a hash on the preceding token. The probability distribution is biased to select green words so that they appear with greater frequency in watermarked text.    
With knowledge of the hash function, we can observe the ratio of green to red words and detect the watermark using a statistical hypothesis test. 
As the text sequence length increases, the probability of misclassifying a text as a false positive is vanishingly small. However, the structured nature of using the preceding token in computing the red-green list introduces vulnerability to perturbation attacks (i.e., making minor modifications or \textit{perturbations} to the text). Word deletions, 
insertions, and synonym replacements can all weaken the watermark signal. As an extreme example, substituting every other word in the passage with a synonym will destroy the watermark completely.
Later \shortciteA{zhao2023provable} simplified the idea to use a fixed red-green list which was shown to improve robustness to perturbation attacks.
The trade-off for enhanced robustness is that an adversary can more easily detect and learn the red-green list, and hence remove or forge the watermark. Note however that an adversary could reconstruct the scheme in either case, given unlimited (black-box) access to the watermark-generating LLM. 

Red-green list methods that alter the relative probability values of the language model are learnable through observation and analysis because they introduce a statistical bias. 
On the other hand, methods that do not alter the probability distribution are said to be \textit{distortion-free}, thereby producing higher quality generations and better imperceptibility. 
Distortion-free methods typically take advantage of the inherent randomness in the token sampling process to evade detection. 
\shortciteA{christ2023undetectable} propose a cryptography-inspired approach in which the decoding process is determined by the value of a secret key, and subsequent detection of the watermark is only possible with knowledge of the key. 
The setup relies on the idea that the language model can be sampled multiple times, each time producing a different valid output drawn from its unaltered probability distribution. Ideally, a randomly generated secret key may be used to select which of these outputs is considered to be watermarked, and the complete process returns that generated text. Without access to the key, the watermark is undetectable because the generated text is just as likely to be observed in the absence of a watermarking intervention. Practically, the secret key is determined using a hash function on the previous text block, as in \shortciteA{kirchenbauer2023watermark}.
\shortciteA{christ2023undetectable} did not provide any experimental results to support their proposed method, but clearly, 
their construction is not robust to text modification attacks since the modified text will likely not map onto the correct secret key value. 
Crucially, \shortciteA{kuditipudi2023robust} introduce the idea of a soft matching function between text and its corresponding key value, and they experimentally demonstrate robustness to paraphrasing attacks in the distortion-free setting.  

For any closed-source model, white-box methods as described above leave all responsibility for watermarking to the owners. 
In contrast, black-box methods take only the generated text as input, enabling anyone to inject a watermark post-generation.
This can be useful for third parties that build applications on top of closed-source language models and want to indicate that text is AI-generated or otherwise embed some copyright information. Early work included formatting strategies such as line shifts \shortcite{Brassil_lineshifts} and Unicode character replacement \shortcite{Rizzo_unicode}, however, these are easily detected and can be removed through text canonicalization \shortcite{liu2023survey}. 
Other methods encode information through syntactic structures \shortcite{topkara_syntax} or vocabulary choice using synonym replacement similar to the idea of red-green lists \shortcite{topkara_synonym}.
In the black box setting, synonym replacement is prone to degrade text quality because the substitution is not selected using the LM's probability distribution. \shortciteA{yang2023watermarking} improves the text quality of black-box lexical-based watermarking by using BERT to select context-aware synonyms. 
Note that many simple watermark techniques are complementary and therefore can be stacked together. For instance, both lexical and syntactic watermarks can be applied simultaneously. The syntactic watermark provides some robustness to synonym replacement attacks that lexical watermarks are vulnerable to. However, both strategies are vulnerable to paraphrase attacks that alter both syntax and word choice. 
One counter idea is to force the adversary to change the most important features of the text (e.g., proper nouns, semantically essential words) to remove the watermark \shortcite{yoo2023robust}, although the quality of watermarked text is also compromised by this approach.
In general, robust black-box watermarking methods seem to be lacking.
A workaround could be to implement white-box methods post-generation by taking the text to watermark, asking an open-source model for a paraphrase, and applying the watermark during the paraphrase generation. Of course, this strategy is only attractive if an open-source model can provide a regeneration of similar quality to the original closed-source generation. 
\shortciteA{zhang2023remark} introduces an end-to-end regeneration strategy (Remark-LLM) that first integrates the watermark into the learned semantic representation, and then uses a trained decoder module to extract the watermark. 
By using a beam search in the decoding step, the regeneration is optimized for coherence and consistency.
Because Remark-LLM is trained end-to-end, the authors include malicious examples in the training data to improve robustness. 
Compared to rule-based approaches, neural watermark embedding requires more text to encode the same amount of information. 


\subsection{Statistical and Stylistic Analysis}
\label{sec:statistical_stylistic}

Both statistical and stylistic analysis involve the extraction of features from the generated text. Statistical methods measure how probable, or likely, each word in the sequence is (relative to some probability distribution). Stylistic methods focus on linguistic properties of the generated text, such as vocabulary, syntax, and coherence.  

\subsubsection{Statistical Analysis}
\label{sec:statistical}

As discussed in Section \ref{sec:background:NLP}, generative language models work by sampling the next word from a learned probability distribution conditioned on the preceding words in the sentence.
Statistical detection methods attempt to recognize any signatures left behind from this sampling process. 
Because knowledge of the probability distribution is generally needed for this approach, most methods in this section can be categorized as \textit{white-box methods}, although recent lines of research attempt to get around this constraint. Additionally, many statistical detection methods are \textit{single-feature classifiers} that use a threshold for class separation. For example, as discussed in Section \ref{sec:background:NLP}, the perplexity (or ``surprisal'') tends to be lower for AIGT than human-written text. Therefore we need only to choose a threshold perplexity value, and we can classify anything below the threshold as AI-generated. 
In the surveyed literature, this category of classifiers is usually called \textit{zero-shot}, implying that no training data is needed. This is likely because the methods are usually evaluated using AUROC, meaning that all threshold values are included in the evaluation, and there is no need for the authors to use training data. However, to use these methods in practice, we do need to choose a threshold, meaning that some training data is needed for threshold calibration. Granted, we could use the published AUROC curves or published feature values to select a threshold without additional training data to use the method in a zero-shot fashion, but this could also be thought of as a pre-trained single-feature classifier. As we will see later in Section \ref{sec:factors:ood}, using new domain- and model-specific calibration data is recommended. Note that having the freedom to choose a threshold means that we can prioritize a low false positive rate, unlike trained classifiers that output a class prediction without an interpretable decision-making process. 

The most trivial implementation of a statistical classifier is to directly measure the likelihood of generating some observed text according to a particular model of interest. 
For each word in a passage, we can use the language model to rank the word choice as well as compute its probability value. Averaging the token-wise probability or rank over the text passage gives us an interpretable score for the text, where higher probabilities and lower ranks indicate AI generation. 
\shortciteA{su2023detectllm} propose using the ratio of average probability and rank since they provide complementary information about the model's certainty of the word choice. The entropy of a text span can also be computed from the probability values to give us an idea of the model's certainty in a word choice given the context. In all cases, these measures are designed to tell us how likely it is that the text was generated by a known model, and some threshold value can be set for classification. 
One advantage of word-level measures is that they allow for intuitive visualization tools (GLTR, \shortciteA{gehrmann2019gltr}) that can help humans understand and recognize AIGT. Another advantage is that they allow for flexible rolling averages over the text passage (i.e., we can detect how measures evolve over a document or concatenation of time-ordered social media posts).  

\begin{figure}[t!]
    \centering
    \includegraphics[width=0.9\textwidth]{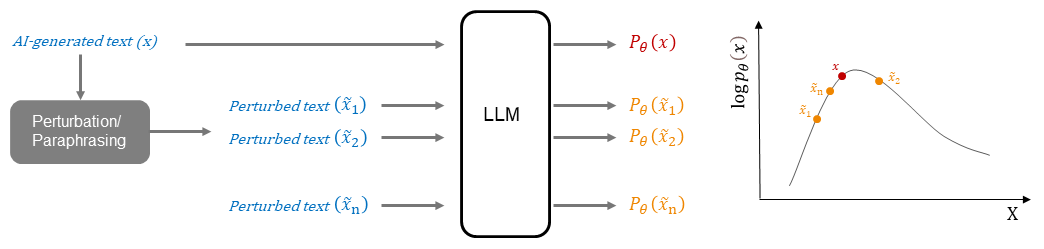}
    \caption{AIGT tends to occupy the negative curvature regions of the probability function \shortcite{mitchell2023detectgpt}.}
    \label{fig:curvature}
\end{figure}

While word-level measures look at the \textit{value} of the language model's probability function in various ways,
we can also learn something from the \textit{structure} of the probability function.
\shortciteA{mitchell2023detectgpt} hypothesize and empirically demonstrate that AIGT tends to occupy the negative curvature regions of the probability function, meaning that any small perturbations of the text will result in lower probability values (see Figure~\ref{fig:curvature}).
Leveraging this observation, they develop DetectGPT, a single-feature method based on a simple computation of the average change in log probability over a sample of text perturbations. While DetectGPT was designed for the white-box, known-model setting, an assumption that we have access to the model that we are trying to detect, experimental results \shortcite{mitchell2023detectgpt,mireshghallah2023smaller} indicate some transferability to the unknown-model setting. \shortciteA{mireshghallah2023smaller} demonstrate that cross-detection can perform almost as well as self-detection, and notably, smaller language models are better universal detectors. 
Also based on the susceptibility of AIGT to local perturbations, \shortciteA{su2023detectllm} define a single-feature measure for the average perturbed rank (NPR). Experimentally, both DetectGPT and NPR achieve impressive detection of GPT2-era generation models in the domain of news and prompted stories.  More recently, \shortciteA{bao2024fastdetectgpt} propose an update to DetectGPT called Fast-DetectGPT, which improves the efficiency of DetectGPT while also increasing the accuracy: experimental results on text from ChatGPT and GPT-4 show Fast-DetectGPT outperforming both DetectGPT and NPR. 
Finally, \shortciteA{venkatraman2023gpt} observe that language models tend to transmit information more uniformly than humans, and therefore suggest the \textit{variance} in word probability over a text span as an informative feature for classification. 

For many practical scenarios, these methods are limited because we cannot access the probability values of closed-source models (e.g., GPT-4 at the time of writing). 
In this case, one way to determine whether text was generated from a particular model is to simply regenerate the text and compare the resulting generation to the original. In this setup, the first portion of the text is retained as the input to the language model, and we do not assume knowledge of the actual prompt that was used.  
\shortciteA{yang2023dna} provide both black-box and white-box variations of this idea, where they compare either output words or probability values, respectively. 
They demonstrate superior detection performance compared to DetectGPT in the white-box setting, and language model-based strategies in the black-box setting. Furthermore, they demonstrate robustness to AI-editing (i.e., perturbations made by another language model). However, the results do depend heavily on the number of regenerations and the truncation ratio (i.e., how much text is used for the input prompt versus the output comparison), and therefore some domain-specific development data may be needed for hyper-parameter selection. 
More specific to the instruction interface of ChatGPT, \shortciteA{yu2023gpt} propose a regeneration method in which ChatGPT both generates the prompt from the original text and then regenerates the output. 
In contrast to language model-based detection methods, the regeneration strategies claim better generalization to unseen data domains and emerging language models, and better robustness to word substitutions, re-translation, and hybrid human-AI texts.  
Note that these methods are most suited to arbitrary or guided generation (i.e., prompts that are not overly specific) in the black-box setting, and it is unclear if they would be successful in detecting controlled generation. 

Another black-box method uses the \textit{intrinsic dimension} of text data as a single-feature threshold method \shortcite{tulchinskii2023intrinsic}. 
The intrinsic dimension of data refers to the minimum number of variables needed to adequately represent that data. It can be thought of as the number of independent underlying variables that are causing some observed data, where the observed data contains noise and is recorded using a higher number of (unknowingly) correlated variables. Dimensionality reduction techniques, such as principal component analysis, aim to extract the true underlying dimension. 
Recall that LLMs represent language as dense ``layers'', which can be thought of as high dimensional noisy data. \shortciteA{tulchinskii2023intrinsic} take the dense representations of all the words in a sentence individually, known as static word embeddings, as the input dataset for dimensionality reduction.
They choose a persistent homology dimension estimator and find that human-written text tends to have an intrinsic dimension between 9 and 10, and AIGT has a lower intrinsic dimension, approximately equal to 8.
This observation seems consistent with the fact that language models tend to produce more unsurprising texts, and therefore adhere to less varied linguistic expressions compared to the full space of possible expressions. Experimentally, the authors show that this separation holds across genres, across generation models (specifically, GPT-2, GPT-3.5, and OPT-13B), and is robust to paraphrasing noise. 
If the observed separation holds universally for all AIGT, regardless of the generating LLM, then this method is applicable for unknown LLMs as well. However, more experimental evidence would be needed to support this claim. 

While regeneration methods and intrinsic dimension estimation do not assume white-box access, they are still most applicable to a known-model detection task. 
Another series of papers proposes using the probability outputs from an ensemble of accessible open-source models only to detect the use of closed-source, unidentified models. The central idea is that human-written text tends to sound equally plausible according to different models, whereas AIGT may be highly preferred by certain models vs.\@ others. 
\shortciteA{li2023origin} extract contrastive probability values between pairs of open-source models which are used as features for a simple linear classifier called Sniffer. By using contrastive features, Sniffer excels in the task of determining which AI model generated the text. 
Similarly, SeqXGPT \shortcite{wang2023seqxgpt} uses probability lists from open-source models as features for a small convolutional network and linear classifier. Unlike Sniffer, SeqXGPT was designed for the task of sentence-level detection within documents with mixed human-AI authorship. In both cases, because the dimension of extracted features is low, only a small amount of training data is needed.
Rather than using the probability vectors directly as features, \shortciteA{verma2023ghostbuster} extract features from probability values using a wider range of open-source models in their ``Ghostbuster'' algorithm. 
Ghostbuster works by passing the input text through a series of weak language models, ranging from a simple unigram model to non-instruction-tuned GPT-3 models (\texttt{ada} and \texttt{davinci}), and computing token probabilities for each token in the input, according to each model. 
Features are extracted from the token probability vectors and a feature selection algorithm is used to obtain the best subset to train a logistic regression classifier. Ghostbuster outperforms existing methods such as DetectGPT and GTPZero on three domains of news, creative writing, and student essays written by GPT-3.5 and Claude. 
In the case that little training data is available, \shortciteA{hans2024spotting} reduce the idea of proxy open-source measures to the single-feature setting. They use cross-entropy between two open-source models, a measure of how surprising a continuation of one model is to another, to detect ChatGPT. Again they find that using a comparative measure between models is more informative than looking at a single model-dependent measure.  
The advantage of this line of research is that it extends detection to new and unseen language models. 

\subsubsection{Stylistic Analysis}
\label{stylistic}

Other methods of detection are based on the idea that AI-generated text has different \textit{stylistic} or \textit{linguistic} properties than human-written text. Some of this research builds on older NLP research areas, such as authorship attribution (determining which human author wrote a document) and stylometry more generally (automated analysis of different literary styles). Because these methods operate on the text-level only, they are suitable in the black-box detection scenario. However, certain stylistic properties might be idiosyncratic to a particular generation model or domain, with varying generalizability to other models and domains.

\shortciteA{frohling2021feature} provide an overview of linguistic features for AIGT detection, along with some discussion of why these features differ between human- and AI-generated text. 
Specifically, they make a distinction between different \textit{decoding} algorithms in AIGT. Recall from Section~\ref{sec:background} that if the decoding method prioritizes the highest-probability words, then the resulting text tends to be fluent but highly repetitive, with low lexical diversity and lack of low-frequency or unusual words. On the other hand, if the decoding method samples from a wider distribution, the resulting text may be low-quality and lacking coherence. In their work, \citeauthor{frohling2021feature} categorize linguistic features according to which potential weakness of AIGT they measure: (1) lack of syntactic and lexical diversity; (2) repetitiveness; (3) lack of coherence; and (4) lack of purpose. They also include a fifth category of general-purpose features, including character-, word-, and sentence-counts, punctuation distribution, and readability metrics. These features are then used to train SVM, logistic regression, random forest, and neural network classifiers. Testing on text generated by GPT-2, GPT-3, and Grover, they find that the syntactic and lexical diversity features, along with the general-purpose features, are most discriminative. The authors note low transferability of classifiers across decoding methods and training sets, suggesting an ensemble method might be more effective than trying to train a single detection classifier to detect all AI-generated text.  \shortciteA{crothers2022adversarial} point to another benefit of an ensemble method, when they find an ensemble of neural and stylistic classifiers is more robust to adversarial attacks at both the word- and character-levels.

Stylistic analysis can also include an investigation into the overall discourse structure of a text. One early work along these lines focuses on the ``factual structure'' of the text \shortcite{zhong2020neural}. That is, do the facts presented in the text flow in a consistent and coherent manner? While effective at the time, it is unclear whether newer LLMs display the same issues with consistency as older models. 
A more recent idea is to look at long-range entity consistency in a text. \shortciteA{liu2022coco} observe that human-written text coherently refers back to previously introduced entities, even with long separations, whereas AIGT tends to cluster same-entity mentions closer together. The entity consistency feature is first extracted as a learned graph structure and then used as input for contrastive learning. Surprisingly, older LLMs are more difficult to detect than newer LLMs using this feature.

\shortciteA{kumarage2023stylometric} focus specifically on the challenging problem of detecting AIGT on Twitter. They propose to use stylistic features to augment baseline methods and overcome the inherent difficulty in classifying such short texts. They additionally tackle the scenario of a tweet sequence being partially authored by humans and partially by AI, aiming to localize where the switch between human and AI occurs. Their methodology involves three categories of features: (1) \textit{phraseology features}, which include measures like the number of words, number of sentences, etc.; (2) \textit{punctuation features}, which involve counting the frequency of occurrence of various punctuation marks; and (3) \textit{linguistic diversity features}, which include measures of lexical diversity and readability. When used alone, the stylistic features are more powerful than some simple baselines based on words alone, but  less discriminative than language model (LM) based classifiers (more information on this family of classifiers in Section~\ref{sec:classification}). However, when the stylometric features are \textit{combined} with the LM-based classifier, the best results are achieved. In a further analysis, the authors find that for the short social media texts in their dataset, punctuation and phraseology features are more informative than the linguistic diversity features, which may require longer texts for accurate estimation. 

In a subsequent work by the same group, \shortciteA{kumarage2023j} present the J-Guard framework to improve the adversarial-robustness of AI-generated news detectors. 
In this work, the method specifically extracts expected features of authentic journalistic text, such as adherence to the AP Style Guide, article structure, avoidance of past tense or passive voice, correct use of punctuation marks, and so on. Adding these stylistic features to a base LM classifier improves performance in all cases, and the J-Guard framework is demonstrated to be better than most existing detectors, while also more robust to paraphrasing and Cyrillic character injection attacks. 

Also working in the domain of news articles, \shortciteA{munoz2023contrasting} compare and contrast the linguistic patterns observed in human-generated text and text generated by the Llama family of LLMs. Using articles from the New York Times as human text, they prompt four different Llama models to generate news articles based on the headline and the first three words of the lead paragraph. Linguistic analysis reveals that the AI generated text has a more restricted vocabulary, uses fewer adjectives and more symbols and numbers, includes fewer words associated with negative emotions, and has a higher propensity for male pronouns. However, the authors do not evaluate the usefulness of these features in a detector, and it is not clear how well these features would generalize across different model families (e.g., GPT) and domains.

Expanding on this idea of domain specificity, many of the discussed methods implicitly assume that human-written text is of higher quality than AI-generated text, using more diverse vocabulary, complex syntax, and demonstrating better coherence. These assumptions are more likely to hold true for older LLMs and in certain domains (such as news articles written by professional journalists). However, in casual writing such as that seen on social media or in text messages, it may actually be the case that AI-generated text can be distinguished by being ``too correct.'' One early work finds that humans are more likely to use clich\'es, idioms, and archaic language than machines, in addition to contractions like \textit{wanna} and \textit{gonna}, and pronunciation spellings like \textit{goin} (for \textit{going})  \shortcite{Nguyen-Son-2017-identifying}. In another human study, participants claim to able to identify AI-generated text because the tone tends to be ``overly formal,'' the statements are too objective and avoid stating any subjective opinions, there is a high frequency of certain phrases such as \textit{it's worth noting} or \textit{please note}, and the texts often involve enumerated lists and end with formal concluding sentences \shortcite{cui2023said}. Thus, it is important to keep in mind the expected level of formality and correctness for human writing \textit{in a given domain} when exploiting stylistic and linguistic properties.

\subsection{Language Model-Based Classification}
\label{sec:classification}


In this section we focus specifically on methods that do not involve an explicit \textit{feature extraction} step, as described in the previous sections, but rather in which the classifier is given the entire text as input and must learn, as part of the training process, which characteristics of the text differ between the classes (AI versus human).
One common classification paradigm in recent years is based on fine-tuning pre-trained language models such as BERT \shortcite{devlin2018bert} or RoBERTa \shortcite{liu2019roberta}. 
As described in Sec~\ref{sec:background:NLP}, these language models are precursors to the current generation of interactive, instruction-tuned LLMs. They have been pre-trained in an unsupervised fashion on large corpora of internet text, which allows them to learn highly effective numerical representations of words and sentences. To leverage these representations for classification, researchers typically add a classification module onto the pre-trained models, and fine-tune the entire model on the domain-specific training data (here, a dataset of AI-generated and human-generated texts). Because of the high dimensionality of the vector representations (typically between 256--768 dimensions, many of which may turn out to be irrelevant to the task), these methods require much more training data than methods which use only a few features. 

This was the approach taken by OpenAI for their first attempt\footnote{OpenAI subsequently trained a more sophisticated model to detect GPT-3 data; however, they later took it offline due to low accuracy \url{https://openai.com/blog/new-ai-classifier-for-indicating-ai-written-text}.} at building a GPT-detector, as described by \shortciteA{solaiman2019release}. They fine-tuned two RoBERTa classifiers (based on RoBERTA\textsubscript{BASE} and RoBERTa\textsubscript{LARGE}), and were able to detect GPT-2 generated text with 95\% accuracy. They also note that the trained classifiers were more accurate than using GPT-2 as a detector itself. They hypothesize that the bidirectional RoBERTa architecture may be more suitable for a discriminative task than the autoregressive GPT, which only takes into account context from the left side. 
From a practical perspective, they also report increased accuracy and robustness when training on a dataset generated using different decoding methods and containing texts of varying lengths.

\shortciteA{chen2023gpt} use a RoBERTa\textsubscript{BASE} model as their pretrained language model, and then fine-tune a multilayer perceptron classifier on top, keeping the RoBERTa weights frozen. They also experiment with a T5 model \shortcite{2020t5}, which they train as a sequence-to-sequence model that outputs either `positive' or `negative', given the text as input \shortcite{chen-etal-2023-token}. They test these two classification models on datasets of text generated by GPT-2 and GPT-3.5 and find that the T5-based model is generally superior; however, both models outperform the OpenAI GPT-detector and zero-shot learning with GPT-2. 

\shortciteA{tian2023multiscale} focus specifically on the problem of classifying short texts, such as SMS texts or tweets, which are very challenging for detection. They propose using a modified ``Positive-Unlabelled'' (PU) framing of the problem, which assumes the training data consists of texts belonging to the ``Positive'' category (i.e., definitely AI-generated) and texts belonging to the ``Unlabelled'' category (could be either human-generated or AI-generated). This captures the intuition that some very short texts, such as \textit{Great weather today!} or \textit{Check this out} cannot be reliably labelled as human or machine generated, and must instead be considered ``unlabelled.'' Using a multi-scale framework to take into account differing text lengths, and adjusting the loss function to allow for unlabelled instances, they finetune BERT and RoBERTa classifiers to achieve state-of-the-art results on short texts in English and Chinese.

The text representations generated by pre-trained language models generally capture aspects of both syntax and semantics \shortcite{jawahar2019does}.  However, \shortciteA{soto2024few} note that when detecting AI-generated text, the \textit{syntactic} component is most relevant, as both humans and AI can generate texts with the same underlying meaning. To this end, they train a BERT-based model to produce \textit{style representations} of text; that is, a vector representation that captures stylistic features rather than semantic features. These style representations are trained entirely on human-generated text from the social media website Reddit. Then, given a small number of calibration samples from the AI model, a test sample is classified as being AI-generated if its style representation vector is ``similar'' (as measured by cosine distance) to the calibration samples. The proposed algorithm performs well experimentally; however, it requires that we know which model generated the text. The method still performs well when extended to multiple AI generators, but only under the assumption that we have access to a sample of training data from each model (i.e., the generating model is not completely unknown).  

As mentioned previously, AIGT detectors in general are vulnerable to paraphrasing attacks, and this also holds true for LM-based classifiers. As a result, one stream of research has focused on improving classifiers' ability to handle paraphrased text. 
\shortciteA{hu2023radar} improve the robustness of a detector to paraphrasing attacks by using \textit{adversarial training} to create the RADAR system (Robust AI-text Detector via Adversarial leaRning). Their training framework involves two competing systems: a Paraphraser, which tries to modify AI-generated text to make it undetectable, and a Detector, which tries to detect the paraphrased AI-generated texts. They choose RoBERTa\textsubscript{LARGE} to initialize the Detector module. The fully-trained detector is shown to be more robust to paraphrasing attacks than DetectGPT or OpenAI's GPT detector. Based on a similar principle, \shortciteA{koike2023outfox} present the OUTFOX framework, which makes use of Attacker and Detector modules. However, in their method, the Detector is based on ChatGPT and uses the adversarially-generated essays for in-context learning to improve detection abilities.

LM-based classification approach has benefits and drawbacks. \shortciteA{li2023deepfake} conduct a series of experiments aiming at recreating a realistic detection scenario, where a detector would encounter a variety of texts from different domains, written by different human authors and an assortment of different AI text generators. In this broad setting, they find that differences in linguistic and statistical features of the texts are less pronounced, and the best-performing method is based on a pre-trained LM called Longformer \shortcite{beltagy2020longformer}. However, it is worth noting that their training set has over 300,000 instances. It may not be feasible to collect training sets of this size (containing both human-written and AI-written examples) in all scenarios. Furthermore, the need to collect new data is constantly ongoing, as research has shown that classifiers trained on text from older models (e.g., GPT-2) do not necessarily perform well for new, larger models \shortcite{ghosal2023towards}.

\subsection{Off-the-Shelf Detection Tools}

In response to the growing need for AIGT detection, a number of different online tools have become available, some of which are summarized in Table~\ref{tab:detection_tools}.

\begin{table}[t!]
    \centering
    \footnotesize
    \begin{tabular}{l l l l}
    \hline
    Name & URL & Language(s) & Min input length \\
    \hline 
    CopyLeaks & \url{copyleaks.com/ai-content-detector}&  30 languages &  350 chars (50-70 words) \\
    GPTKit & \url{gptkit.ai} & English & 50 words\\
    GPTZero & \url{gptzero.me} & English &  250 chars (40-50 words)\\
    Originality AI &\url{originality.ai/ai-checker}  & 16 languages & 50 words\\
    Sapling & \url{sapling.ai/ai-content-detector}& English & 50 words \\
    Scribbr & \url{scribbr.com/ai-detector} & 3 languages & 25 words \\
    Winston AI&\url{gowinston.ai} & 6 languages & 600 chars (90-120 words) \\
    Writer & \url{writer.com/ai-content-detector} & English & 60 words\\
    ZeroGPT &\url{zerogpt.com}  & 6 languages & 450 chars (70-90 words) \\
    \hline
    \end{tabular}
    \caption{Some of the available off-the-shelf detection tools. }
    \label{tab:detection_tools}
\end{table}

Because these tools are commercial products, they have released varying amounts of information about their underlying algorithms. GPTZero, for example, is based on the academic research of one of its co-founders, and so somewhat more information about its method is available \shortcite{tian2023identifying}. One of the core features it measures is ``burstiness,'' a measure of the variability in style, tone, and vocabulary throughout a document (presumed to be higher in human-written texts). It also makes use of a proprietary LLM to compute probabilities for each word, given a context. Additionally, it searches the web to determine whether a text input already exists, and finally also has a deep learning component trained on massive datasets of human-written text and text from various language models. Similarly, GPTKit uses an ensemble of six classifiers, based on both statistical measures as well as pre-trained language models \shortcite{orenstrakh2023detecting}. 
It is likely that other successful commercial tools also make use of multiple different components to improve the robustness and reliability of the outputs, rather than relying on a single underlying methodology. 

Comparing the performance of the different tools is challenging, as it depends on both the test data and the evaluation metric. In July 2023, \shortciteA{orenstrakh2023detecting} compared several different online tools for detecting AIGT specifically in the domain of computer science education. They found that CopyLeaks was the most accurate detector, while GPTKit had the best false positive rate. In August 2023, GPTZero released data\footnote{\url{https://gptzero.me/news/gptzero-surpasses-competitors-in-accuracies}} showing very high performance for its own model as well as CopyLeaks, and lower accuracy by Originality.ai and ZeroGPT. In September 2023, \shortciteA{akram2023empirical} performed a different evaluation, finding that Originality.ai had the best recall and precision overall for both human and AI texts (0.96--0.98), with GPTZero, Sapling, and Writer considerably lower, and GPTKit classifying almost everything as human-written. In short, there is no clear ``winner'' among the available tools, and submitting a text to multiple tools and aggregating the results may give a more reliable result than any one tool itself.

\subsection{Humans as Detectors}

Finally, to underscore the difficulty of this task, we summarize the research findings on how well humans can detect AIGT. A number of studies show that humans cannot reliably distinguish between AI- and human-written text with higher accuracy than that achieved by simply guessing randomly. 

\shortciteA{liu2023check} find that human performance is close to random guessing on a AIGT detection task involving scientific abstracts, and that human annotators have the tendency to think that all abstracts are human-written. \shortciteA{sarvazyan2023overview} also report that human annotators perform at near the random guessing baseline. While language proficiency can be a factor, they observe no significant difference for annotators experienced with AIGT versus those with no prior exposure. 
Similarly, \shortciteA{li2023deepfake} ask three annotators majoring in linguistics to try to distinguish between AI- and human-generated text; 
their performance is also only slightly better than chance. The creators of the Ghostbuster algorithm ask 6 undergraduate and PhD students with familiarity with AIGT to attempt the detection task, resulting in an average accuracy of 59\% \shortcite{verma2023ghostbuster}.

Some studies report better results using different methodologies; for example, \shortciteA{guo2023close} find that it is easier for humans to detect ChatGPT when provided with a pair of responses (one human, one ChatGPT). In this context, experts familiar with ChatGPT are much more successful than amateurs.  In contrast to that result, \shortciteA{uchendu2021turingbench} report that humans detect AIGT at chance level, both when shown single instances \textit{and} in the paired `pick the AI' task.  In the domain of student essays, \shortciteA{liu2023argugpt} find that ESL teachers can detect AI-generated essays with an accuracy of 61\%, and this can be improved to 67\% with minimal exposure and self-training. 

These human studies motivate the importance of developing computational tools for the AIGT detection task. The question is not simply one of scale: the capabilities of these models to generate realistic, fluent text has exceeded our human ability to detect it as computer-generated. Computational detection algorithms are able to outperform human annotators due to their ability to analyze various statistical properties of the texts that are not immediately perceptible to human observers.

\subsection{Summary of Existing Methods}

\begin{figure}[tbh]
    \centering
    \includegraphics[width=0.9\textwidth]{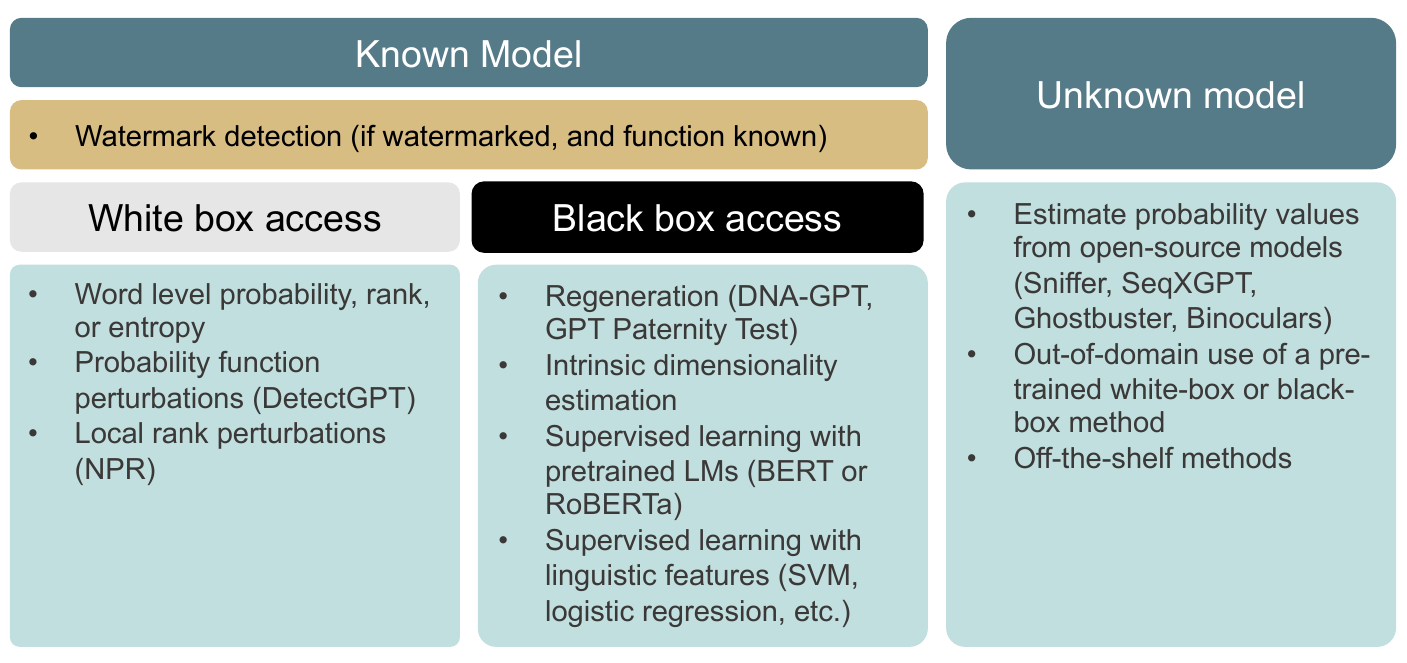}
    \caption{Detection methods available in different detection scenarios.}
    \label{fig:detection_scenarios}
\end{figure}

The choice of which detection method to use, out of the many that have been presented, will depend on a number of factors. In Section~\ref{sec:background:detection_scenarios} we outlined the most common detection scenarios; we now revisit these categories in Figure~\ref{fig:detection_scenarios} together with the different approaches presented in this section. 

The ability to use watermark detection as a strategy is dependent, of course, on whether the text has been watermarked in the first place. In this scenario, only access to the watermark extraction function is needed, and not access to the generating model itself; however, it is assumed that the model is known. 

In cases where the model is known but not watermarked, and we have access to the internal parameters, methods based on the probability, perplexity, and entropy of individual words or sequences can be used to help detect text generated by that model. If, on the other hand,  the unwatermarked model is known but we have only black-box access, regeneration strategies can be used instead. Another standard approach for the known model, black-box access scenario is supervised learning, either by first extracting linguistic/stylistic features and training a low-dimensional classifier, or fine-tuning a high-dimensional, pre-trained language model based on the raw texts. 

If the generating model is not known, then we must use known models as proxies. This can involve either estimating probability values based on open-source models, or using classifiers (or off-the-shelf tools) trained on other models. In these cases, we are relying on general properties of AIGT that are expected to hold true \textit{regardless} of which specific model generated the text. However, experimental results suggest that the detection performance of most methods is somewhat lower when applied to text from an unknown model.

In practice, there are numerous other factors which can affect the performance of the detection methods outlined above. 
These factors are discussed in detail in 
Section~\ref{sec:factors}. However, we first summarize the available datasets for AIGT detection. 

\section{AIGT Datasets}
\label{sec:datasets}


\begin{table}[!t]
\centering
\footnotesize
\begin{tabular}{lllr}
\toprule
                                      \textbf{ Dataset Name} &         \textbf{Language} &                \textbf{Domain} & \textbf{Dataset Size} \\
\midrule
    tum-nlp/IDMGSP \shortcite{abdalla2023benchmark} &          English & Academic publications &          24K \\
                 GPABench2 \shortcite{liu2023check} &          English & Academic publications &        2800K \\
                      CHEAT \shortcite{yu2023cheat} &          English & Academic publications &          50K \\
                 OUTFOX \shortcite{koike2023outfox} &          English &                Essays &          30K \\
                 ArguGPT \shortcite{liu2023argugpt} &          English &                Essays &           8K \\
                    LLMFake \shortcite{chen2023can} &          English &        Misinformation &           5K \\
           COVID-19 Tweets \shortcite{choi2024fact} &          English &        Misinformation &           4K \\
              PropaNEWS \shortcite{huang2022faking} &          English &        Misinformation &           2K \\
          Fake News \shortcite{huang2023harnessing} &          Chinese &        Misinformation &           40K \\
      Fake News \shortcite{jiang2023disinformation} &          English &        Misinformation &          50K \\
                      News \shortcite{kreps2022all} &          English &        Misinformation &           1K \\
                   F3 \shortcite{lucas2023fighting} &          English &        Misinformation &          40K \\
                       ODQA \shortcite{pan2023risk} &          English &        Misinformation &          25K \\
           News \shortcite{schuster2020limitations} &          English &        Misinformation &           2K \\
         Real/Fake Tweets \shortcite{spitale2023ai} &          English &        Misinformation &         0.2K \\
                 GossipCop++ \shortcite{su2023fake} &          English &        Misinformation &          20K \\
                PolitiFact++ \shortcite{su2023fake} &          English &        Misinformation &         0.5K \\
News and Social Media \shortcite{zhou2023synthetic} &          English &        Misinformation &         0.5K \\
             RAID \shortcite{dugan-etal-2024-raid} &          English &      Multiple domains &         6200K \\
                      MixSet \shortcite{gao2024llm} &          English &      Multiple domains &           4K \\
                MGTBench \shortcite{he2023mgtbench} &          English &      Multiple domains &          18K \\
              In-the-wild \shortcite{li2023deepfake} &          English &      Multiple domains &         447K \\
              SnifferBench \shortcite{li2023origin} &          English &      Multiple domains &          36K \\
  AuTexTification \shortcite{sarvazyan2023overview} & English, Spanish &      Multiple domains &         160K \\
                       HC3-SI \shortcite{su2023hc3} & English, Chinese &      Multiple domains &         215K \\
       Ghostbuster \shortcite{verma2023ghostbuster} &          English &      Multiple domains &          21K \\
                          M4 \shortcite{wang2023m4} &     Multilingual &      Multiple domains &         150K \\
          SeqXGPT-Bench \shortcite{wang2023seqxgpt} &          English &      Multiple domains &          36K \\
            HC-Var \shortcite{xu2023generalization} &          English &      Multiple domains &         145K \\
             CT2 \shortcite{chakraborty2023counter} &          English &                  News &        1600K \\
    MULTITuDE \shortcite{macko-etal-2023-multitude} &     Multilingual &                  News &          74K \\
     TuringBench \shortcite{uchendu2021turingbench} &          English &                  News &         200K \\
 FakeNews/RealNews \shortcite{zellers2019defending} &          English &                  News &          25K \\
                       HC3 \shortcite{guo2023close} & English, Chinese &    Question-answering &         125K \\
              OpenOrca \shortcite{Lian2023OpenOrca} &          English &    Question-answering &        4200K \\
           TweepFake \shortcite{Fagni2021TweepFake} &          English &          Social Media &          25K \\
       GPT-wiki-intro \shortcite{aaditya_bhat_2023} &          English &              Websites &         300K \\
                OpenLLMText \shortcite{chen-etal-2023-token} &          English &              Websites &          30K \\
\bottomrule
\end{tabular}
\caption{Commonly used datasets with AI-generated texts.}
\label{tab:datasets}
\end{table}

As described in the previous section, there are three main methods for the detection of AI-generated text (AIGT): watermarking, statistical and stylistic analysis, and the use of pre-trained language models (LMs). The detection of a watermark requires knowledge of the watermark extraction algorithm; beyond that, no extra data are needed. However, the other two broad detection methods require data in order to learn the patterns that distinguish AIGT from human-written text -- ideally, a dataset where the positive examples were generated by the AI model we wish to detect, and in a setting as close as possible to how we expect to encounter these texts in the real world. Previous work has indicated that the most effective detectors are trained on data from the same \textit{domain} (news articles, social media posts, academic essays, etc.), \textit{language} (English, Chinese, French, etc.) and \textit{model settings} (decoding algorithm, prompt, length of output, etc.) as the test data. At the same time, the research also shows that for maximal generalizability and robustness, it is essential to train a detector on a wide variety of data so that it is not overfit to one very narrow range of data samples.  Therefore, for any given application, it is important to select appropriate data to first train the detector, and also to test the detector for its accuracy: ``The effectiveness of black-box detection models is heavily
dependent on the quality and diversity of the acquired data'' \shortcite{tang2023science}.

Table~\ref{tab:datasets} lists some of the most frequently used datasets that include human-written and AI-generated texts. 
We observe that the vast majority of available datasets are in English, with another sizeable chunk being multilingual (including English as well as other languages).  There is a well-known bias in the field of NLP that much of the research focuses on English, to the detriment of other languages. This bias is no doubt exacerbated by the fact that many of the large language models (LLMs) used to generate these datasets were, at least initially, only available in English. However, we expect the situation to continue to change as more and more multilingual LLMs become available. Among the languages other than English, Chinese is represented the most. 

As mentioned above, another important factor is the \textit{domain} of the dataset. We observe the 
presence of expected domains such as news, academic writing, and essays. Social media, while highly interesting from a practical perspective, is not well-represented, potentially due to the difficulty of detecting AIGT in short texts, or of simulating realistic social media posts for dataset creation. 
Some datasets that focus more on casual, online settings with multiple authors include: HC3 \shortcite{guo2023close}, M4 \shortcite{wang2023m4}, In-the-wild \shortcite{li2023deepfake}, AuTexTification \shortcite{sarvazyan2023overview}, MixSet \shortcite{gao2024llm}, OpenLLMText \shortcite{chen-etal-2023-token}, HC-Var \shortcite{xu2023generalization}, Real/Fake Tweets by \shortciteA{spitale2023ai}, and COVID-19 Related AI-Generated Tweets by \shortciteA{choi2024fact}. 
A fairly large proportion of the datasets include multiple domains. For example, the M4 dataset \shortcite{wang2023m4} is a large multi-purpose dataset containing seven different languages, and multiple domains and generators, including state-of-the-art LLMS like GPT-4. Similarly, the RAID benchmark \shortcite{dugan-etal-2024-raid} contains generations from eight different domains, 11 models, and four different decoding strategies. 
Not surprisingly, many datasets with AI-generated content were created for the task of misinformation detection. We further note that within this broad category, there are multiple sub-categories (e.g., fake news, social media misinformation, etc.).  

Dataset size is computed as the total size by combining the human-generated and AI-generated examples. This is a rough characterization since certain datasets may be imbalanced in favour of one or the other. Furthermore, some datasets contain a very large number of samples from a single model or language, while other datasets comprise a smaller number of samples from a large number of different models. However, we see an overall trend for datasets composed of fewer than 50,000 samples.
While some detection methods claim to be zero-shot or few-shot methods, requiring fewer data samples to calibrate the algorithm, it is generally considered beneficial to have as much data as possible.  


It is important to note that, with very few exceptions, all of these datasets were \textit{generated} for research purposes, as opposed to \textit{collected} from online sources. The reasons for this are self-evident: if we don't have an accurate AIGT detector in the first place, we cannot determine whether any given text on the internet has been written by a human or by AI. By generating the AIGT themselves, researchers can guarantee that it is in fact AIGT, and by limiting their human data samples to those written before, say, 2020, they can be reasonably confident that they were actually written by humans. One notable exception is the TweepFake dataset \shortcite{Fagni2021TweepFake}, which includes data scraped from known bot accounts on Twitter for the AIGT portion of the dataset. 

Instead, most datasets are generated by first defining a ``parent'' or ``anchor'' dataset of human-generated text, and then artificially generating parallel AIGT text. For example, in the news domain, researchers might start with a corpus of news stories. They then feed the headline (or the headline plus the first sentence) into an LLM and ask the LLM to write the rest of the article. In this fashion, they can generate a parallel dataset of human- and AI-written articles for the same set of headlines. Similarly, in the domain of question-answering, if the researcher has a dataset of questions with human-written answers, they can ask the AI to answer the same questions. 
Some generation methods include more specific style prompting to mimic the human dataset, as in ``\textit{Write a news article in the style of the New York Times}.''  
A more challenging detection scenario is \textit{controlled generation}, where the LLM is more constrained by the parameters of an existing human-written text, as in the case of summarization or paraphrasing. Datasets that tackle this issue include CHEAT \shortcite{yu2023cheat}, GPABench2 \shortcite{liu2023check}, tum-nlp/IDMGSP \shortcite{abdalla2023benchmark}, HC3-SI \shortcite{su2023hc3}, and OpenLLMText \shortcite{chen-etal-2023-token}. An even more challenging scenario for AIGT detection is \textit{mixcase}, or documents that contain a mix of human and AI writing. Some of the datasets that include this type of data are: MixSet \shortcite{gao2024llm}, GPAbench2 \shortcite{liu2023check}, and SeqXGPT-Bench \shortcite{wang2023seqxgpt}. 
Such mixcase scenarios are especially critical in the domain of misinformation detection where a documents containing a mixture of true and false statements pose a significant challenge for detection algorithms. Such scenarios include: perturbing just a few facts in real human-written articles, as in F3 \shortcite{lucas2023fighting} and ODQA \shortcite{pan2023risk}, removing and adding negations, as in AI-Generated News by \shortciteA{schuster2020limitations}, finding and replacing the most salient sentence in the story, as in  PropaNEWS \shortcite{huang2022faking}, or mixing a real and a false story in one article, as in Human-Written and AI-Generated Fake News by \shortciteA{jiang2023disinformation}. Further, generating and editing a story in a multiple-step dialog with the model, as done in e.g., AI-Generated Fake News by \shortciteA{huang2023harnessing} and Human-Written and AI-Generated Fake News by \shortciteA{jiang2023disinformation}, may produce even more sophisticated fake news, but would require more human resources.  
All these methods of generating text allow for precise, controlled research experiments on detecting AIGT; however, it is unclear how well they represent the artificially generated text that actually exists on the internet. 

Datasets with a parallel structure (that is, the human and AI samples were generated from the same prompt or question) can be valuable to ensure topic consistency between the human and AI subsets. However, as mentioned, it is preferable for the AIGT to have been generated from multiple prompts so that the detector does not learn spurious correlations related to the specifics of a given prompt. Some datasets that include parallel data with multiple prompting strategies include: HC-Var \shortcite{xu2023generalization}, In-the-wild \shortcite{li2023deepfake}, and Ghostbuster \shortcite{verma2023ghostbuster}.

Another related factor to consider is that the detector is not only learning a model of AIGT from the training data, but also a model of what \textit{human} writing looks like. So if, for example, the training data only includes samples from professional journalists, as is the case in many news datasets, we cannot expect the detector to accurately recognize text written by the average lay-person. A lack of diversity in the human-generated samples is likely the cause of biases such as the observation that the AIGT detectors can mis-classify writing by English language learners as being AIGT. Therefore, the training data should ideally represent a diverse variety of text from \textit{both} AI and humans.

\section{Factors that Influence Detectability}
\label{sec:factors}

There are numerous aspects that influence the difficulty of the AIGT detection task. These can include properties of the generating model, such as size and decoding method, properties of the text, such as length and language, properties of the training data, such as domain and prompt design, and human influence factors, such as mixed human-AI authorship and purposeful adversarial evasion. The findings give insight into the limitations of the detection methods and highlight important considerations for compiling training data when building a detection system.  

\subsection{Properties of the Generating LLM}
\label{sec:factors:LLM}

Some features of AIGT that affect detectability can be traced back to the properties of the generating model. 
Specifically, the size and decoding strategy of the generating LLM can significantly affect the detectability of the generated text. 

Generally speaking, increasing the size (i.e., number of trained parameters) of a language model is one way to improve the quality of the generated text. Unsurprisingly then, larger language models (e.g., GPT-4) are more difficult to detect. 
\shortciteA{chakraborty2023counter} introduce the AI Detectability Index (ADI), an experimentally motivated index to quantify the detectability of LLMs. Based on their observation that newer, larger language models have statistical signatures (e.g., perplexity and negative-log curvature signals) approaching that of human distributions, they assign higher ADI to larger models. 
\shortciteA{pagnoni2022threat} show that this property holds true even when the \textit{same} generating model is fine-tuned for detection (i.e., the detector has the same capacity for representing language as the generator). Experimentally, they observe that detectability is related to model size by a power law, meaning that detection accuracy decreases linearly as the number of model parameters increases exponentially. 

Secondly, the decoding strategy of the generating model affects detectability. 
Recall from Section~\ref{sec:background:NLP} that a decoding strategy refers to the method of choosing the next word to generate, given the probability distribution over possible choices. Top-$k$ sampling selects among a fixed number of choices, whereas nucleus sampling allows more possible choices when the model is uncertain. 
\shortciteA{pagnoni2022threat} observe that detection accuracy decreases significantly when training data is generated using a different decoding strategy than test data (e.g., a decrease of 21\% is observed when the detector is trained on top-$k$ decoding output and tested on nucleus sampling output compared to in-distribution results). In general, nucleus sampling output is the most difficult to detect, and detectors trained on nucleus-sampled data also generalize the best across other sampling methods.  
\shortciteA{pu2023deepfake} show that even within the same decoding strategy, changing the decoding parameters between training and inference time can significantly impact detectability. For instance, when using nucleus sampling, the recall of a statistical detector on AIGT drops by 13\% by changing the probability threshold from 0.96 to 0.8. Likewise, for top-$k$ sampling, recall falls by 56.4\% by changing the $k$ from 40 to 160.  
\shortciteA{stiff2022detecting} also highlight the potential difficulty of detecting text output by a generative discriminator decoding strategy. In this setup, a smaller auxiliary language model is used to guide the larger language model toward generating text with a specific attribute (e.g., positive sentiment). Most detection methods do not seem to consider the possibility of generative discriminator decoders. 

\subsection{Language}
\label{sec:factors:language}

In most of the studies surveyed, it is assumed that the text to classify is written in English, and training data is also written in English.  
\shortciteA{wang2023m4} investigate using XLM-RoBERTa, a cross-lingual version of RoBERTa, to detect AIGT in multi-lingual settings. The model is trained by seeing examples in one language, and needs to generalize detection to multiple unseen languages. This ability is useful for low-resource languages where ample training data is not available. However, the results show that cross-lingual detection is a challenging task. The detection methods generally struggle, though there is some limited ability to detect Chinese when trained on English data and vice versa.
Because cross-lingual training and detection is an unsolved challenge, AIGT in low-resource languages has limited detectability. The M4 dataset published by \shortciteA{wang2023m4} can be used to benchmark future work on cross-lingual generalization.  That said, a number of the off-the-shelf tools in Table~\ref{tab:detection_tools} claim to work in multiple languages, mostly high-resources languages such as French, Spanish, German, Chinese, etc. Presumably, these tools train separate, monolingual detection models for each language.

A different language consideration pertaining to human-written text is English as a second language.
\shortciteA{liang2023gpt} observe that texts written by non-native English speakers are more likely to be falsely classified as AI.
Using seven publicly available detection tools, including GPTZero and ZeroGPT, they find that detection accuracy is almost perfect on US 8th-grade student essays whereas TOEFL essays written by Chinese English learners have a false positive rate close to 60\%. \shortciteA{ardito2023contra} explains that secondary language learners are more likely to use common vocabulary and adhere to simple syntax and style formulas, similar to language models that tend towards unsurprising sentences. The result is that detection methods are generally biased against those with less varied linguistic expression.    
Using this observation, \shortciteA{liang2023gpt} experiment with bias mitigation and evasion techniques by asking ChatGPT to alter text in a specific way. Whether applied to human-written text or AIGT, they find that the prompt ``Enhance the word choices to sound more like a native English speaker'' produces an output that is more likely to be classified as human-written.  

\subsection{Document Length}
\label{sec:factors:length}

Naturally, shorter text sequences contain less information and are therefore more difficult to classify. Watermarking methods purposefully inject algorithmically detectable information and require only on the order of 10 words to carry a binary signal (signifying AIGT or not). Statistical, stylistic, and finetuning-based methods work to detect statistical patterns in the existing text and require on the order of at least 100 words for accurate classification. Experimentally, \shortciteA{li2023deepfake} observe that approximately 120 words are sufficient for both statistical (GLTR, DetectGPT) and finetuning-based classifiers to reach their full potential across a wide range of task domains (story generation, news writing, and scientific writing). In the same ballpark, \shortciteA{he2023mgtbench} find that approximately 200 words are sufficient to detect powerful LLMs such as ChatGPT-turbo and GPT-4.    
\shortciteA{chakraborty2023counter} study the question of sequence length from a theoretical perspective and prove that it is always possible to detect AIGT given an adequate sequence length, except when the AI and human text distributions are identical. They show that even if the total variation distance between distributions is very small, an assumption likely to occur as language models become more powerful, a sequence length of around 500 words should be sufficient.
If text sequences do not meet these length requirements, as might be common for social media posts, it is still possible to improve detection by concatenating disjoint posts from the same author. \shortciteA{stiff2022detecting} improve the accuracy of a finetuned RoBERTa detector from 80\% to near 100\% using a concatenation of 10 tweets. In this case, even if the individual text samples don't form a coherent continuation, the increased sequenced length obtained through concatenation improves detectability.    

Note that the dependence on text length is not independent of the generator's decoding strategy. As discussed above, nucleus sampling produces text that is harder to detect, likely because the text becomes less deterministic when the generating model is uncertain. This means that text generated by nucleus sampling is more susceptible to length requirements. \shortciteA{pagnoni2022threat} show that shortening the text length from 256 words to 64 words produces a 10\% drop in accuracy on texts produced by nucleus sampling, compared to less than 1\% on texts produced by top-k sampling.  

Besides the amount of information content, the sequence length of samples in the training data also impacts the required text length at inference time. 
\shortciteA{xu2023generalization} observe that text generated from ChatGPT tends to be longer than human-written text given the same prompts. If synthetic training data is not controlled for length, the detector is biased by the text length and tends to misclassify longer human-written text as AIGT.
Similarly, \shortciteA{guo2023close} find that shorter sentences are harder to detect if they are not seen enough in the training data. Conversely, if the training data includes short sentences, then the model can generalize to longer texts. \shortciteA{liu2023argugpt} also observed within their ArguGPT benchmark that sentence-level detection cannot be achieved if the training set consists only of full essays. They found that by adding sentence-level examples to the training data, accuracy on sentence-level detection increased from 50\% (random guessing) to 93\%. If sentence-level detection within a longer document is important, then it seems essential that synthetic training data also includes shorter sentence-length examples.   

\subsection{Out-of-Distribution Domains and Generating Models}
\label{sec:factors:ood}
When classifiers are trained or calibrated on a certain type of data, and then expected to make inferences on different, unseen types of data, this is said to be \textit{out-of-distribution}. In the context of detectability, many robustness studies look at whether detectors can generalize to unseen data domains (e.g., news, social media, etc.), unseen generating models, and unseen generation prompts.  

In general, adapting to mixed or unseen data domains is a challenging task for existing detection methods. 
\shortciteA{pu2023deepfake} focus on ``real-world'' synthetic text and publish four new datasets from three commercial generation-as-a-service providers and a ChatGPT-powered bot on Reddit. They find that existing methods (including statistical metrics and finetuned classifiers) are not as strong on the ``in-the-wild'' datasets compared to their original claims using synthetic research datasets. 
\shortciteA{li2023deepfake} observe that stylistic and statistical (GLTR, DetectGPT) methods fail when the test data contains multiple data domains, even if all domains were seen during training (detection accuracy drops by 10-25\%), and detectability is further impacted by unseen data domains (dropping accuracy by another 1-15\%). Language model-based methods are 
able to generalize to multiple data sources, but they are not immune to unseen data. The accuracy of a finetuned LongFormer detector \shortcite{beltagy2020longformer} drops by over 20\% in the out-of-
distribution setting, compared to multi-domain performance. 
\shortciteA{wang2023m4} also conduct a cross-domain study of existing methods (GLTR, GPTZero, and finetuned RoBERTa) and find that RoBERTa generalizes the worst despite having the strongest in-domain performance.
\shortciteA{pagnoni2022threat} hypothesize that perhaps RoBERTa's training objective is less robust to finetuning for AIGT compared to other language models, such as ELECTRA. 
RoBERTa uses a masked language modeling objective, whereas ELECTRA uses a discriminator to identify spans of text that were replaced by a language model.
Experimentally, ELECTRA is shown to be more robust to unseen data.   
\shortciteA{he2023mgtbench} note that some data domains are more transferable than others, observing that language model-based methods are more flexible when trained on creative writing data with many human authors rather than news data with few human authors. 
Despite the poor generalization capability of language model-based detectors, \shortciteA{xu2023generalization} show that transfer learning is possible, meaning that existing trained detectors can be adapted using a small amount of new domain data. 

In the case of unseen generating models, we observe that statistical detectors generalize poorly while finetuned language model-based detectors generalize to some extent, according to performance on several benchmark datasets including
ArguGPT \shortcite{liu2023argugpt}, MGTBench \shortcite{he2023mgtbench}, and M4 \shortcite{wang2023m4}. 
Notably, GPTZero is not able to detect essays written by a model other than GPT-3.5 for which it is calibrated \shortcite{liu2023argugpt}, and DetectGPT does not transfer between ChatGPT and GPT-3, two closely related language models \shortcite{abdalla2023benchmark}. 
Although language model-based detectors might generalize better than statistical detectors, their performance is still shown to be unsatisfactory in some contexts \shortcite{he2023mgtbench,abdalla2023benchmark}. 
Furthermore, their generalization ability is conditional on the input text length and size of the generating model. 
\shortciteA{liu2023argugpt} observe that RoBERTa's accuracy drops by only 2\% on full essay examples, but drops by 13\% on sentence length texts, compared to in-distribution performance. 
\shortciteA{pagnoni2022threat} show that smaller LLMs struggle to detect text produced by larger LLMs, although this disparity becomes less pronounced among the largest models. 
Contradictory results across evaluation studies may speak to this conditional generalizability of language model-based detectors.  

Finally, out-of-distribution data may occur when the generating \textit{prompt} differs from training data to test data. For example, generating stories via the prompt ``tell me a bedtime story about $\langle$context$\rangle$'' will produce a different distribution than ``write an exciting short story about $\langle$context$\rangle$'', though they might be closely related. In some sense, using different prompts can be thought of as out-of-distribution tasks. \shortciteA{liu2023argugpt} experiment with essay generation prompts that optionally contain additional instructions such as ``use specific reasons and examples to support your answer''. 
Note that \shortciteA{liu2023argugpt}'s prompt construction does not contain anything adversarial or unexpected for essay generation, but they find that both GPTZero and fine-tuned RoBERTa still have some difficulty in generalizing to the unseen prompt.  \shortciteA{xu2023generalization} find that RoBERTa generalizes between highly similar prompts well in some domains (news, movie reviews) and less so in others (creative writing, question answering). Furthermore, they show that prompts that produce responses that are closer to human responses 
(as measured by MAUVE score,  \shortciteA{pillutla-etal:mauve:neurips2021}) 
create training data that promotes generalizability. When human-written training data is available, it may be worthwhile to prompt-tune AI-written examples to obtain human alignment.  
Generalizing between highly similar prompts is necessary because, though we might reasonably have knowledge of the general task and data domain, we cannot assume knowledge of the specific prompt wording. 

\subsection{Degree of Human Influence}
\label{sec:factors:human_influence}

As discussed in Section~\ref{sec:taxonomy}, the degree of human influence on AIGT can vary wildly, and this also greatly impacts detectability. 
\shortciteA{liu2023check} evaluate detectors' ability to distinguish between scientific abstracts that were either human-written, GPT-written, GPT-completed, or GPT-polished. 
The GPT-polished category contains abstracts that were fully written by humans and then rewritten by ChatGPT for clarity. As expected, GPT-polished text is the hardest to detect; the evaluated methods (including GPTZero, ZeroGPT, and OpenAI's detector) perform worse than random guessing.  
Similar to \shortciteA{liu2019roberta}'s GPT-polished category, \shortciteA{abdalla2023benchmark} 
evaluate detection performance on scientific papers that were human-written, and then paraphrased by ChatGPT.
They obtain a detection accuracy of 75\% on this dataset by finetuning RoBERTa with access to human-AI paraphrased data during training. 
Although some human effort is required to write the original text in full under this scenario, we can imagine that multiple varied copies could be created by using language models to paraphrase in different styles. In the context of misinformation, this could be used to lend credibility to an idea being posted by multiple authors. 
Furthermore, the ability to control the style of the generated text, for example by asking a model to re-write a given text in the style of a reputable source, poses a significant challenge to detection tools \shortcite{wu2023fake}.

\shortciteA{gao2024llm} present results on their MIXSET dataset, which contains human-written text revised by AI under the categories of: 
i) \textit{polish}: the AI makes word and sentence level substitutions for clarity, ii) \textit{complete}: the AI completes the remaining 2/3 of a document, and iii) \textit{rewrite}: the AI extracts ideas from a human-written text and completely rewrites the full text. Flipping the roles, the dataset also contains AIGT that is post-edited by humans under categories of i) \textit{humanize}: add typos, grammatical errors, etc. (note this is simulated using AI) and ii) \textit{adapt}: humans rewrite the text for fluent and natural-sounding language. Of the statistical (\shortciteA{gehrmann2019gltr,verma2023ghostbuster,mitchell2023detectgpt}) and finetuned (\shortciteA{hu2023radar,guo2023close,chen2023gpt}) detection methods included in the experiments, none are able to perform well at binary AIGT detection if mixcase examples are not seen during training. Of the mixcase categories, AI paraphrasing and humanizing appear to be the most difficult, while human polishing (adapt) seems the least detrimental to detectability. The finetuned detectors seem especially brittle to typos, and threshold-based detectors are especially brittle to AI paraphrasing. In general, AI polishing at the sentence level worsens detectability more than polishing at the word level. However when mixcase examples are seen during training, the Radar detector \shortcite{hu2023radar} achieves approximately 88\% detection accuracy across all mixcase categories.

Similarly, human-AI cooperation can present additional challenges for misinformation detection. 
As described in Section~\ref{sec:datasets}, 
an LLM can be asked to insert some false information in otherwise trustworthy content, for example by merging two articles, a real and a fake ones, slightly modifying some facts, or inserting a fake fact into a real story \shortcite{schuster2020limitations,huang2022faking,jiang2023disinformation,lucas2023fighting}. Psychological studies on human deception confirm that the hardest lies to detect are the ones that are closest to the truth, i.e., where only a small amount of facts are altered and it is done in a minimal way \shortcite{mazar2008dishonesty}. In such cases AI-generated misinformation becomes very difficult to detect by both humans and AI \shortcite{schuster2020limitations}. 

Another type of human-AI hybrid text is machine-translated text. \shortciteA{weber2023testing} take human-written text in seven languages (Bosnian, Czech, German, Latvian, Slovak, Spanish, and Swedish) and use Google Translate to generate the English translation. Using 14 publicly available online tools including DetectGPT, GPTZero, and OpenAI's text classifier, they find that approximately 95\% of the original texts are correctly classified as human-written. After machine translation, around 70\% of texts are classified as human-written. 
Note that the sample size of documents in this study is too low to draw any conclusive insights. 

\subsection{Adversarial Attacks}
\label{sec:adversarial}

The previous subsections summarize detection difficulties that are prone to arise naturally. In this section, we turn our attention to adversarial attacks, a scenario in which an adversary purposefully attempts to evade detection. 
Most adversarial attacks alter AIGT using either word substitutions or paraphrasing, meaning there is some overlap with the ideas in the previous sections (i.e., an adversary can exploit poor generalization ability). 
Most adversarial attacks work under the restriction that the semantic meaning of the original text should not be altered by the attack, but some loosen this restriction.  
Borrowing language from \shortciteA{shi2023red}, a language model is said to be \textit{protected} if a detector exists that can detect language ordinarily produced by the language model. Some adversarial attacks assume access to an unprotected model to launch their attack, while some work under the more challenging setting that any auxiliary language model used in the attack is protected.  

\shortciteA{pu2023deepfake} develop a word perturbation attack based on knowledge of the detection method and the generating model. That is, they focus on detection methods that were designed or calibrated to detect a particular white-box generation model, and then alter AIGT from that model to evade detection. 
Their attack works by searching for words in the text that are most important for classification and then switching those words for low-probability synonyms that maintain the overall text quality and semantics.  
Specifically, they select the most confidently predicted words according to the generating model by word rank, ignoring stop words. Potential synonym replacements are then selected using the cosine similarity of the word embeddings, and the part-of-speech tags are also matched to preserve the grammatical structure. Finally, they check the similarity between the original and perturbed sentence representations to verify that the overall meaning is maintained. Of the candidate synonyms meeting these criteria, they choose the one with the lowest rank according to the generation model to make the substitution.
Note that this method requires access to the generating model (which is assumed to be protected), but does not require access to another unprotected language model. 
Detection methods that rely on probability-based metrics (e.g., average token-wise probability, rank, and GLTR) are easy to fool with this attack, but a stylistic method that focuses on the global factual structure of text \shortcite{zhong2020neural} is robust. 
\shortciteA{wang2024stumbling} introduce a different type of word- and character-level attack by introducing typos and character homoglyphs. Again, statistical classifiers are shown to be brittle to this attack, but we note that the attack is easily removed by pre-processing at inference time. 

Beyond word-level attacks, paraphrasing attacks alter AIGT more aggressively by rewriting the entire text.  
\shortciteA{krishna2023paraphrasing} build their own paraphrasing model (DIPPER) and demonstrate that input from three LLMs (GPT2, OPT-13B, GPT-3.5) can be paraphrased to evade detection by watermarking \shortcite{kirchenbauer2023watermark}, GPTZero, DetectGPT and OpenAI's classifier.
Robustness is reported using the true positive rate (TPR) at a fixed false positive rate of 1\%, and the statistical detectors are shown to be the most brittle at less than 5\% TPR, followed by the finetuned classifier at 13\%, followed by watermarking at 50\%.  
DIPPER was designed to produce semantically faithful paraphrases and does not assume query access to the detection methods (i.e., the paraphrases are not adversarially optimized to avoid detection). 
Note that this attack requires the use of an \textit{external} paraphraser (essentially an unseen LLM that is not protected). Viewed this way, the paraphrasing attack just exploits the lack of transferability of known-model detection methods. 
Building on DIPPER as the base paraphrasing model, \shortciteA{sadasivan2023can} show that iterative paraphrasing is effective in further breaking the detection methods. The evasion rates even out after 5-6 iterations and text quality is degraded only slightly as measured by perplexity, performance on test benchmarks, and human evaluation.

The above paraphrasing methods work under the assumption that the paraphrasing LLM is not protected (i.e., the existing detection methods are not trained to detect it), however, most detection methods could be adapted to detect the new model (e.g., a language model-based classifier can be finetuned on the paraphrased output, and there's no reason to believe that this a more difficult task).  
\shortciteA{shi2023red} focus on the more challenging and realistic scenario in which any auxiliary LLM used for the attack is also assumed to be protected. They propose two types of attacks, one based on word substitutions, and one based on paraphrasing. 
Unlike \shortciteA{pu2023deepfake}, word replacements are selected by an auxiliary LLM to provide more contextualized synonyms. 
In the query-free setting, words are selected to replace at random\footnote{Except in the case of watermarking when the highest entropy words are selected. The distribution bias introduced by red-green list methods \shortcite{kirchenbauer2023watermark} affects high entropy words (those with the most uncertainty) the most.}, and in the query-based setting, an evolutionary search algorithm is used to select evasive perturbations. Note that query-based attacks are optimized to evade a particular detector. 

\shortciteA{shi2023red}'s paraphrasing attack uses prompt tuning to generate texts that are difficult to detect. Notably, the new paraphrase is not restricted to be semantically similar to the original generation, but it must be an equally valid generation given the original prompt. Without access to repeated queries to the detector and adaptive prompt tuning, a RoBERTa-based detector actually is robust to paraphrasing (expected as it is trained to detect both the generator and paraphrasing model). However, with prompt tuning, the detection method can successfully be evaded. 
Note that \shortciteA{shi2023red} use a type of adversarial finetuning based on only the binary output from a detector, but with access to the detector's classification ``score'', even more techniques would become available. 
For example, \shortciteA{henrique2023stochastic} propose reinforcement learning with a trained detector as the critic model.  


While we commonly think of adversarial attacks as avoiding the AIGT classification, some attacks actually work to forge the AIGT label. This is mostly specific to watermarking methods, where the presence of a watermark is meant to certify the authorship of the text.    
\shortciteA{sadasivan2023can} show that red-green list watermarking methods are susceptible to spoofing attacks that allow a human to forge the watermark. A malicious actor could write offensive language while injecting a watermark to damage the reputation of the LLM owner. As discussed in Section~\ref{sec:watermarking}, watermarking methods that introduce a bias in the text distribution can be inferred by repeated black-box sampling only. Experimentally, \shortciteA{sadasivan2023can} find that around 1 million queries are needed to sufficiently predict the red-green labels of the 181 most common words in the English language.

\section{Discussion and Summary}\label{sec:discussion}

As we have seen, many factors affect the difficulty of AIGT detection, ranging from the intuitive (e.g., larger LLMs are more difficult to detect) to the surprising (e.g., it can be hard to generalize between similar prompts). 
Taken together, these research findings offer insight into the strengths and limitations of existing detection methods. When implementing a solution for a particular application, it will be necessary to balance multiple considerations. Based on the literature, we offer a few high-level recommendations.


Some researchers argue that as LLMs approach human-level generation capability, watermarking methods will be the only viable approach to detection \shortcite{tang2023science}.
On 7 February 2024, OpenAI announced the use of a watermark in images created using DALL-E 3 and ChatGPT\footnote{https://help.openai.com/en/articles/8912793-c2pa-in-dall-e-3} with a publicly available detection interface. However, text watermarking remains a challenge, and we cannot assume that AIGT gathered online from unknown sources will contain a watermark. Therefore, while watermarking is an active research area and an important aspect of AIGT detection, it will not likely play a role in current detection strategies until there is greater industry adoption.

Classifiers based on the statistical properties of the generated text are intuitively logical, as they attempt to capture regularities in the text that occur as a direct result of the generation process. However, accurate knowledge of the probability function requires white-box access to the generating LLM, which is often not available. In such cases, methods that use open-source LLMs as proxies have been shown to be effective, in particular when multiple LLMs are used together. Stylistic and linguistic properties of AIGT have the benefit of being measurable without any access to or knowledge of the generating LLM; however, they may vary widely across models and domains. When a large quantity of labelled training data is available, pretrained LMs perform well \shortcite{li2023deepfake}. However, this necessitates an ongoing process of data collection as models trained on older generating models do not generalize well to newer models.

Although many of the findings in Section~\ref{sec:factors} suggest that statistical metric-based classifiers are more brittle than LM-based classifiers when generalizing to unseen detection scenarios, it is likely that statistical methods will still have a place in an ensemble detection strategy. This is because threshold-based methods can be calibrated to have low false positive rates in their specialized detection setting. That is, a statistical method can be calibrated to catch any AIGT that falls into a narrow specification, such as a certain data domain, generating model, and even generation parameters, reliably without misclassifying many instances of human-written text as AI. A large ensemble of differently calibrated metrics for known detection cases, along with the more generally applicable language model-based classifiers for unknown detection cases, will most likely be needed for robust AIGT detection.

Another key component in AIGT detection is choosing appropriate and realistic training data. 
 A good match to the target language and domain is important. If there is a particular generating LLM that is of most interest, finding a large dataset generated by the same LLM will be critical. However, given the easy availability of LLMs, it is quite feasible for a researcher to develop their own dataset for a particular project. The biggest challenge in this task is generating AIGT that is ecologically valid; that is, which imitates to a high degree of accuracy the AIGT data that will actually be observed ``in-the-wild.'' Based on the literature, we summarize the following high-level recommendations for generating training data:

\begin{itemize}

    \item Generate data in the language of interest. If multiple languages are of interest, generate separate monolingual datasets to train separate detectors. Several LLMs now generate text in multiple languages. Previous research has found cross-lingual AIGT detection performance to be poor, though this may change as the field advances \shortcite{wang2023m4}. 
    \item When creating the dataset, include examples (of both human-written and AIGT) of variable length including sentence-level examples. Check that the length distribution is balanced between the two classes.  In some cases, this can be achieved by including a length variable in the prompt \shortcite{liu2023argugpt,xu2023generalization}. If the AI model consistently generates text that is longer than human text, the trained detector will perform poorly for short AIGT, or long human-written text \shortcite{xu2023generalization}.
    \item If the data domain is known, generate in-domain data for the training set \shortcite{liu2023argugpt}. Otherwise, generate mixed-domain data for maximal generalizability. 
    \item Include human-written examples by many different authors, ideally with different writing styles and degree of language proficiency \shortcite{tang2023science}.  
    \item Include AIGT samples from a variety of prompts, not just a single prompt \shortcite{xu2023generalization,liu2023argugpt}.
    \item Where possible, tune prompts to produce output that is similar to the human-written training examples \shortcite{xu2023generalization}, for example using the MAUVE similarity score \shortcite{pillutla-etal:mauve:neurips2021}. Alternatively, filter out AIGT samples that are not sufficiently human-like \shortcite{pagnoni2022threat}.
    \item If it is necessary to detect mixed human-AI authorship cases, be sure to include examples of this in the training data \shortcite{gao2024llm}.
    \item Where possible, generate AIGT using nucleus sampling as the decoding strategy with a range of different hyper-parameter values \shortcite{pagnoni2022threat,tang2023science}.
\end{itemize}

\section{Challenges and Future Directions}\label{sec:conclusion}

In this survey we have described the existing NLP methods for the detection of AI-generated text, as well as some of the evasive tactics that can be deployed to avoid detection. While there is no single method that can be recommended in all scenarios, there are a number of different promising options depending on the information and resources available. Additionally, several research papers as well as the existing commercial tools point to the utility of ensemble methods, where multiple detection strategies are deployed in combination to improve the robustness of the overall detection.  However, as LLMs continue to grow in size and complexity, their output has become harder and harder to detect, and there is no reason to think that this trend will not continue into the future. Therefore, there are some key challenges that must be addressed in future work. 

One important research direction will involve combining NLP techniques with other methodologies for more robust AIGT detection. These complementary methods could include human domain experts, particularly in specialized areas such as second-language learning, misinformation detection, and sensitive sectors such as military applications. As one successful example of this, \shortciteA{gehrmann2019gltr} found that when human judges were provided with a graphical interface that combined automated probability information along with the text, their accuracy at detecting AIGT increased from 54\% to 72\%.  Injecting domain expertise into detection systems, as well as incorporating human-in-the-loop methodologies, will therefore both be important directions to explore. Additionally, the use of non-linguistic properties to detect AIGT in certain situations, such as on social media, can also be used in combination with the NLP methods. For example, the related problem of \textit{social bot detection} is often approached by taking into account posting frequency, network characteristics, and profile information \shortcite{stieglitz2017social,cresci2020decade}.  Generative AI is also being increasingly used to generate multimedia content, such as images or videos \shortcite{lin2024detecting}. Analysing multimedia content in conjunction with text may also provide informative results.  

Another area for work is in properly calibrating uncertainty estimations for each classification decision. Rather than simply outputting a binary ``AI'' or ``human'' label, it is more actionable and useful to know with what level of certainty that determination is being made. Particularly in high-stakes applications for human writers, such as academic coursework or professional writing, being falsely accused of using AI can have serious negative consequences. Knowing if the detector is 55\% certain or 99\% certain in its analysis can help inform what action to take in each case. Previous work in text classification has documented the fact that language model-based classifiers are often unjustly confident in their predictions, and accurately calibrating uncertainty estimations remains challenging \shortcite{kong2020calibrated}. Accurate certainty estimations can also be beneficial in human-in-the-loop scenarios, where high-certainty decisions are handled by the machine but uncertain cases are sent for a more detailed human review. 

Up to this point in time, major players like OpenAI and Meta have dominated the LLM market, and the massive resources required to train an LLM from scratch means that most users have relied on one of the available models, keeping the set of possible generating models to a relatively small size. However, some researchers have raised concerns about the availability of open-source models and the potential for users to retrain and modify such models to make them essentially ``unknown'' to existing detectors \shortcite{tang2023science}. 
New tools such as low-rank adaptation (LoRA, \shortciteA{hu2021lora}), retrieval augmented generation (RAG, \shortciteA{lewis2020retrieval}), and OpenAI's GPTs\footnote{\url{https://openai.com/blog/introducing-gpts}} make it increasingly easy and affordable to adapt LLMs to specialized datasets.
It is currently unclear how these adaptations change the statistical features of the generated text, i.e., the features that are leveraged by existing detectors. 
Testing detectors on 
fine-tuned, personalized,  or otherwise modified LLMs will be another new frontier in this area.


There have been calls for regulation and legislation around the release of generative AI tools. In one proposal, \shortciteA{knott2023generative} call for legislation requiring every company that releases a generative AI model to also release a public-facing detector for content generated by that model. They suggest that this could be achieved with watermarking, or by simply keeping a private log of all text (or other media) generated by the model and then searching through the records to determine whether a given test instance is in the log. While acknowledging that this will not stop bad actors from developing their own private models, it would trivialize detection of text from any of the larger, commercially available models. Whether or not such legislation will be passed, and in what regions, remains to be seen.

Most studies on detecting AI-generated text in general, and AI-generated misinformation in particular, have focused primarily on the domains of news articles and academic writing -- both relatively long and formal genres of text. However, it is clear that social media posts are also powerful vectors for the spread of misinformation. Research suggests that not only are social media platforms targeted by organized political propaganda and disinformation campaigns, but that the algorithms underlying the platforms have a tendency to promote sensational and attention-catching content, whether it is true or not \shortcite{wardle2017information}. Therefore, it will be essential that future work investigates the detection of AI-generated misinformation on social media, as well as news media.

Finally, there is increasing awareness in the field of AI generally that models should be evaluated on metrics other than task performance in order to ensure their safety and adherence to ethical standards, and this no doubt applies to AIGT detectors as well. In Section~\ref{sec:factors:language} we highlighted the findings of \shortciteA{liang2023gpt}, who found that AIGT detectors demonstrated a bias against non-native English speakers. Related to this, certain communities may be more likely to rely on generative AI for non-malicious purposes, such as translation or proofreading, and should not suffer unnecessary consequences for this. Thus, fairness and non-discrimination are important desiderata for any automated system. Governmental guidelines on automated decision-making in Canada\footnote{\url{https://www.tbs-sct.canada.ca/pol/doc-eng.aspx?id=32592}} and the European Union\footnote{\url{https://ec.europa.eu/newsroom/article29/items/612053}} also emphasize the importance of \textit{explainability}, or having a model be able to output a meaningful explanation for its decisions. This has been largely overlooked in the existing literature, and deserves further consideration. 

The rise of generative AI in the past few years has represented a paradigm shift in computer science, and in society more broadly. The use of this technology is widespread and at this point cannot reasonably be contained, and therefore we must adapt our ways of thinking to co-exist with the use of generative AI. Society is now in the process of determining what co-existence looks like, as we are still in the early stages of understanding both the negative and positive consequences of this technology. For example, teachers have had to modify the kinds of assignments they create to avoid plagiarism, while also taking advantage of the new opportunities for interactive and personalized learning. \shortciteA{ardito2023contra} argues that AIGT detection has no place in educational settings at all, and that learning objectives should instead shift to cooperative use of AI. Similarly, LLMs have unfortunately proven to be highly effective generators of misinformation, but researchers have also designed new ways to leverage LLMs to help users with fact-checking and information verification. 
Therefore, it is clearly not the case that all AI-generated text is `bad' or generated with malicious intent, just as it is evident that not all human-generated text is `good,' factual, or helpful. Nonetheless, the integrity and trustworthiness of our information ecosystem depend on being able to reliably determine the source of information, whether human or AI, in order to properly assess its credibility. Consequently, the problem of automated AI-generated text detection is imperative at present, and will continue to be so in the future.







\section*{Acknowledgments}
This project was funded by the Canadian Department of National Defence, Annex Number: ER-VAL 23-085. The authors thank Zachary Devereaux for insightful feedback and discussion.



\vskip 0.2in
\bibliography{sample,bibliography,bibliography_datasets}
\bibliographystyle{theapa}

\end{document}